\documentclass[sigconf, nonacm]{acmart}

\usepackage{amsmath,amsfonts}
\usepackage{algorithmic}
\usepackage{graphicx}
\usepackage{textcomp}
\usepackage{times}   
\usepackage{helvet}   
\usepackage{courier}   
\usepackage[ruled,norelsize,vlined]{algorithm2e}
\usepackage{multirow}
\usepackage{multicol}
\usepackage{subfig}
\usepackage{marvosym}
\begin{document}

\title{DWCL: Dual-Weighted Contrastive Learning for Multi-View Clustering}

\thanks{*Equal Contribution \quad \textsuperscript{+} Corresponding author}
\thanks{This work is supported by the National Natural Science Foundation of China (No. 42371480, 62306033) and the Beijing Natural Science Foundation (No. L247008).}

\author{Hanning Yuan$^{1}$*, Zhihui Zhang$^1$*,  Lianhua Chi$^3$,  Qi Guo$^1$}
\author{Sijie Ruan$^{1}$, Jinhui Pang$^{1+}$, Xiaoshuai Hao$^{2+}$}
\affiliation{%
  \institution{\textit{$^1$Beijing Institute of Technology, Beijing, China}\\ \textit{$^2$Beijing Academy of Artificial Intelligence, Beijing, China}\\
        \textit{$^3$La Trobe University, Australia}\\
}
}
\email{{3220231441,yhn6,3220241531,sjruan,pangjinhui}@bit.edu.cn}
\email{L.Chi@latrobe.edu.au }
\email{xshao@baai.ac.cn}

\begin{abstract}
Multi-view contrastive clustering (MVCC) has gained significant attention for generating consistent clustering structures from multiple views through contrastive learning. However, most existing MVCC methods create cross-views by combining any two views, leading to a high volume of unreliable pairs. Furthermore, these approaches often overlook discrepancies in multi-view representations, resulting in representation degeneration. To address these challenges, we introduce a novel model called Dual-Weighted Contrastive Learning (DWCL) for Multi-View Clustering. Specifically, to reduce the impact of unreliable cross-views, we introduce an innovative Best-Other (B-O) contrastive mechanism that enhances the representation of individual views at a low computational cost. Furthermore, we develop a dual weighting strategy that combines a view quality weight, reflecting the quality of each view, with a view discrepancy weight. This approach effectively mitigates representation degeneration by downplaying cross-views that are both low in quality and high in discrepancy. We theoretically validate the efficiency of the B-O contrastive mechanism and the effectiveness of the dual weighting strategy. Extensive experiments demonstrate that DWCL outperforms previous methods across eight multi-view datasets, showcasing superior performance and robustness in MVCC. Specifically, our method achieves absolute accuracy improvements of 3.5\% and 4.4\% compared to state-of-the-art methods on the Caltech5V7 and CIFAR10 datasets, respectively. 

\end{abstract}

\maketitle


\begin{figure}[h]
    \centering
    \includegraphics[width=0.95\linewidth]{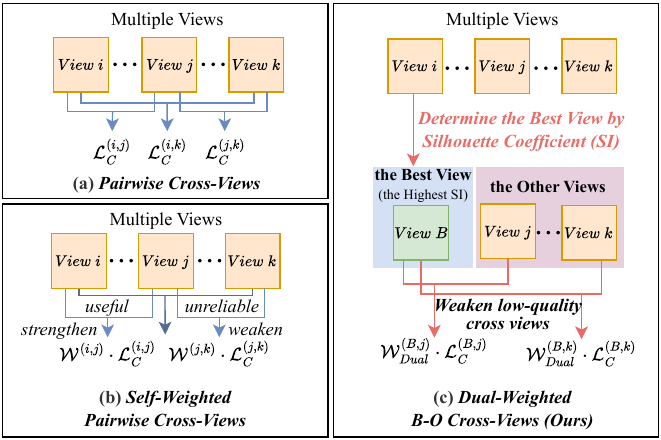}
    \Description{}
    \caption{
    Comparison of our DWCL with existing works in cross-view construction. 
(a) Pairwise cross-view construction often results in a large number of unreliable cross-views, which limits the representation of individual views.
(b) Self-weighted cross-view construction, based on view discrepancy, can unintentionally amplify low-quality cross-views that exhibit low discrepancy in contrastive learning.
(c) In contrast, our approach employs the silhouette coefficient to create dual-weighted cross-views, ensuring that the representation learning of individual views is guided by the highest-quality view.}

    \label{motivation}
      \vspace{-0.8em}
\end{figure}

\section{Introduction}
Multi-view clustering (MVC) enhances clustering quality by leveraging complementary information from multiple views. It has garnered significant attention across diverse fields, including image and video representation~\cite{sun2021scalable,yu2024dvsai}, social network analytics~\cite{cui2021mvgan,zhang2024multiview}, and cross-media information retrieval~\cite{jin2023deep,xu2023adaptive,wang2023self}.
Contrastive learning plays a crucial role in this context by explicitly enhancing the similarity between representations of semantically related instances. This paradigm strengthens representations across multiple views and is increasingly adopted in MVC~\cite{trosten2023effects,lin2022dual,xu2022multi,yang2023dealmvc}. 
Multi-view contrastive clustering (MVCC) methods focus on constructing cross-views to effectively learn representations from multi-view data, where cross-view refers to the involvement of any two views in the contrastive learning process~\cite{chen2023deep}.

Most traditional multi-view contrastive clustering (MVCC) methods~\cite{yang2023dealmvc,huang2024generalized,cui2024novel} utilize a pairwise cross-view construction method, as shown in Fig.~\ref{motivation}(a), treating all cross-views formed by any two views as equally important in contrastive learning. 
However, views in multi-view data inherently vary in quality, being either strong or weak~\cite{yang2021deep}. This pairwise approach can lead to cross-views of differing quality, and the representation similarity between high-quality and low-quality views can converge towards 1.0 in contrastive learning, thereby constraining the representation capabilities of individual views within low-quality cross-views. Additionally, the prevalence of unreliable cross-views increases computational costs. Moreover, representation degeneration resulting from view discrepancy remains a significant challenge in MVCC~\cite{wang2022contrastive,wen2023scalable,huang2023self,cui2024dual,yan2024anchor}. While some strategies, such as the self-weighting method proposed in SEM~\cite{xu2024self}, as shown in Fig.~\ref{motivation}(b), aim to mitigate this issue by applying a view discrepancy weight to enhance low-discrepancy cross-views and diminish high-discrepancy ones, they often overlook view quality, inadvertently promoting low-discrepancy but low-quality cross-views. This highlights the need for a more balanced approach that considers both view quality and discrepancy.

\begin{figure*}
    \centering
\includegraphics[width=1\linewidth]{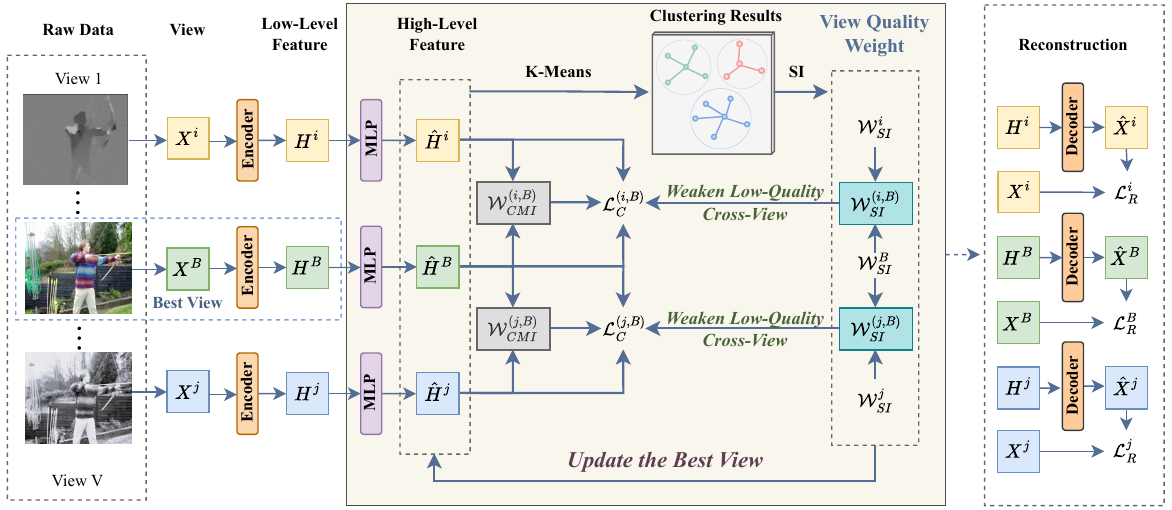}
    \Description{}
    \caption{
The framework of DWCL. Silhouette coefficient (SI), the internal evaluation index of k-means clustering, is utilized to determine the best view. In our Best-Other (B-O) contrastive mechanism, specific cross-views are formed by combining the best view with other views. The view quality weight, $\mathcal{W}{SI}$, is integrated with the view discrepancy weight, $\mathcal{W}{CMI}$, to effectively reduce the influence of low-quality and high-discrepancy cross-views in the dual-weighted contrastive learning.
    }
    \label{fig:DWCL}
\end{figure*}

To address the challenges of pairwise cross-view construction and self-weighting strategies, we propose a novel model called Dual-Weighted Contrastive Learning (DWCL) for Multi-View Clustering, as illustrated in Fig.~\ref{fig:DWCL}. This model introduces an innovative multi-view contrastive mechanism known as the Best-Other (B-O) contrastive mechanism, which aims to enhance the representation capabilities of individual views while reducing the computational burden of numerous unreliable cross-views in MVCC. 
In the B-O contrastive mechanism, cross-views consist of the best view identified by the silhouette coefficient (SI)~\cite{rousseeuw1987silhouettes} paired with the other views. The complexity of the B-O mechanism is reduced from quadratic to linear, $i.e.$, from $O(|V|^2)$ to $O(|V|)$, where $V$ is the number of multiple views, compared with the pairwise contrastive mechanism. Additionally, we propose a view quality weight, $\mathcal{W}_{SI}$, that utilizes SI to evaluate quality discrepancies among views. To mitigate representation degeneration, we incorporate a dual weighting strategy within the B-O contrastive mechanism, combining the view quality weight with a view discrepancy weight, $\mathcal{W}_{CMI}$~\cite{xu2024self}, as shown in Fig.~\ref{motivation}(c). This strategy enhances high-quality and low-discrepancy cross-views while downplaying low-quality and high-discrepancy ones. Extensive experiments demonstrate that DWCL outperforms existing MVCC methods across eight multi-view datasets. 

\textbf{In summary, our main contributions are as follows:}
\begin{itemize}

\item 
In this paper, we propose Dual-Weighted Contrastive Learning (DWCL) model to effectively and efficiently address representation degeneration in Multi-View Contrastive Clustering (MVCC).

\item We introduce an innovative and efficient Best-Other (B-O) contrastive mechanism, the first to enhance the representation ability of individual views while significantly reducing computational costs associated with numerous unreliable cross-views.

\item We design a novel dual-weighting strategy that incorporates a view quality weight to assess view reliability, combined with a view discrepancy weight to address cross-view inconsistencies. This strategy effectively strengthens high-quality with low-discrepancy cross-views while suppressing low-quality with high-discrepancy ones, ensuring robust multi-view representation learning.

\item We provide rigorous theoretical justifications and proofs to  validate the efficiency of the B-O contrastive mechanism and the effectiveness of the dual weighting strategy.

\item 
Extensive experiments on eight multi-view datasets demonstrate the effectiveness of the proposed DWCL. Specifically, our method achieves a 3.5\% absolute improvement in accuracy compared to state-of-the-art methods on the Caltech5V7 dataset, and a 4.4\% absolute improvement on the CIFAR10 dataset. 

\end{itemize}

\section{Related Work}
\subsection{Deep multi-view clustering} 
Recently, multi-view clustering (MVC)~\cite{wan2023auto,wan2024one,wan2024fast} has made significant progress. Leveraging the robust representational capabilities of deep networks, deep MVC has been extensively applied in various fields. Deep MVC is generally divided into two categories: two-stage algorithms and one-stage algorithms. Two-stage MVC, as explored in works such as~\cite{li2019deep,xu2021multi}, first learns feature representations from different views and then performs clustering. In contrast, one-stage MVC, as detailed in~\cite{zhou2020end,lin2021completer}, integrates the extraction of intrinsic structures from multi-view data with clustering into an end-to-end framework. These methods utilize deep neural networks for multi-view clustering, effectively uncovering latent clustering patterns within the data. However, challenges remain in optimizing their performance across diverse datasets.

\subsection{Contrastive learning} 
Contrastive learning has emerged as a powerful paradigm for unsupervised and self-supervised representation learning. Its fundamental principle is to bring similar instances closer in the representation space while pushing dissimilar instances apart. Pioneering methods such as SimCLR~\cite{chen2020simple} and MoCo~\cite{he2020momentum} focus on learning instance-level representations by leveraging data augmentation and contrasting positive and negative pairs. These approaches emphasize the importance of data augmentation strategies and the design of contrastive loss functions, such as the InfoNCE~\cite{oord2018representation} loss, in achieving effective representation learning.
Recent advancements have aimed to address the limitations of traditional contrastive learning methods. For instance, SwAV~\cite{caron2020unsupervised} expands the contrastive framework by integrating clustering techniques to improve representation consistency. However, significant challenges persist due to inherent distributional differences and feature space inconsistencies in multi-view scenarios, making it difficult to balance the need for consistency across views while preserving the unique characteristics of individual views.

\subsection{Multi-view contrastive clustering} 
In multi-view learning, contrastive learning is commonly used to align representations across different views while preserving their unique information. Multi-view contrastive clustering (MVCC) has advanced significantly in self-supervised representation learning. Traditional methods~\cite{lin2022dual,tang2022deep,tang2022DSI,xu2022multi,chen2023deep} focus on learning consistent representations for the same instances by constructing pairwise cross-views. For example, DualMVC~\cite{yang2023dealmvc} aligns global view features with pseudo-labels to capture consistency among these features.
However, real-world multi-view data~\cite{winn2005locus,lin2012human} often introduces semantic discrepancies between views, resulting in cross-view differences that do not arise in traditional contrastive learning, where cross-views are typically generated through data augmentation. These discrepancies present significant challenges in MVCC.
MVCC methods depend on generating numerous cross-views for effective representation learning, but many existing approaches~\cite{ma2024cross,pan2021multi,trosten2023effects,hao2023mixgen,huang2024generalized,cui2024novel} combine any two distinct views, leading to an excess of unreliable pairs that can degrade the quality of individual view representations. While methods like CVCL~\cite{chen2023deep} enhance clustering quality by comparing assignments across multiple views, and SEM~\cite{xu2024self} proposes a self-weighted MVCC approach to mitigate representation degeneration from discrepancies, few address the varying quality of views. The challenge remains to reduce the impact of noisy or low-quality views, as these can significantly impair overall representation quality.
In contrast, our approach introduces the Best-Other (B-O) contrastive mechanism, which constructs reliable cross-views by incorporating both view quality and view discrepancy weights. This contrastive mechanism adaptively adjusts the contribution of each view, helping to address representation degeneration and effectively manage the challenges posed by
low-quality views.

\section{Method}

\subsection{Problem Statement}

Given a set of multi-view data $\{\boldsymbol X^v \in R^{N \times D_v}\}_{v=1}^V$ with $V$ views and $N$ instances of $D_v$ dimensions, $\boldsymbol X^v = [\boldsymbol x^v_1, ..., \boldsymbol x^v_i, ...,\boldsymbol x^v_N]$ represents the $v$-th view and $\boldsymbol x^v_i$ donates the $i$-th instance of the multi-view data. 
Instances with the same semantic label can be categorized into the same cluster. 
Assume $K$ is the number of clusters, the goal of MVC is to partition $N$ instances into $K$ distinct clusters according to their semantic similarity.

\subsection{Determine the Best View in the Best-Other (B-O) Contrastive Mechanism} 
We propose a Best-Other (B-O) contrastive mechanism, in which the cross-views consist of the best view and any other views.
Given the multi-view data $\{\boldsymbol X^v \in R^{N \times D_v}\}_{v=1}^V$, firstly, a low-level representation feature $\{{\boldsymbol H}^v\in R^{N \times {H}_v}\}_{v=1}^V $ for each view is obtained by the $v$-th encoder, represented as:
\begin{equation}
{\boldsymbol H}^v = f_E^v(\boldsymbol X^v;\theta^v).
\end{equation}
In the low-level representation feature space, view-specific auto-encoder modules learn discriminable features across multiple views. To further explore the consistency information shared among the views, we employ high-level representation features for contrastive learning. 
The high-level representation feature $\{\hat{\boldsymbol H}^v\in R^{N \times \hat{H}_v}\}_{v=1}^V $ for each view is obtained by employing the $v$-th MLP on ${\boldsymbol H}^v$, expressed as:
\begin{equation}
\hat{\boldsymbol H}^v=f_M^v({\boldsymbol H}^v;\phi^v), 
\end{equation}
where $\theta^v$ and $\phi ^v$ are network parameters. 
Then, we employ k-means~\cite{macqueen1967some} to derive the cluster labels of instances in each view. For the $v$-th view, the cluster label of the $i$-th instance $\hat{\boldsymbol y}_i^v$, is obtained by:
\begin{equation}
    \hat{\boldsymbol y}_i^v = \mathop{\arg\min}\limits_{k}\Vert \hat{\boldsymbol h}_i^v - {c}_k^v \Vert_2^2,
    \label{klabel}
\end{equation}
where ${\hat{\boldsymbol h}^v_i\in \hat{\boldsymbol H}^v}$, ${\hat{\boldsymbol y}^v_i\in \hat{\boldsymbol y}^v}$, ${c}_k^v$ is the centroid of cluster $k$ for the $v$ view, and $\{{c}_k^v\}_{k=1}^K$ denote the $K$ cluster centroids.

The Silhouette Coefficient (SI)~\cite{rousseeuw1987silhouettes}, a common clustering evaluation metric, is used to assess the clustering quality of each view in the multi-view scenario. The clustering result for each view is obtained through k-means, and the view with the highest SI is selected as the best view for that dataset. The SI for the $v$-th view, denoted as $SI(\hat{\boldsymbol H}^v)$, is computed as the average Silhouette Coefficient of all instances in the representation $\hat{\boldsymbol H}^v$, as given by:
\begin{equation}
SI(\hat{\boldsymbol H}^v)=\frac{1}{N}\sum_{\hat{\boldsymbol h}^v_i\in \hat{\boldsymbol H}^v}SI(\hat{\boldsymbol h}^v_i),
\end{equation}
where:
\begin{flalign}
      &  SI(\hat{\boldsymbol h}^v_i) = \frac{b(\hat{\boldsymbol h}^v_i) - a(\hat{\boldsymbol h}^v_i)}{\max\{b(\hat{\boldsymbol h}^v_i), a(\hat{\boldsymbol h}^v_i)\}},\label{eq_SI}\\
 &   a(\hat{\boldsymbol h}^v_i) = \frac{1}{|C_k^v|}\sum\nolimits_{\hat{\boldsymbol h}^v_j \in C_k^v, \hat{\boldsymbol h}^v_i \neq \hat{\boldsymbol h}^v_j} d(\hat{\boldsymbol h}^v_i,\hat{\boldsymbol h}^v_j),\\
&    b(\hat{\boldsymbol h}^v_i) = \min_{l\in [1,K],l\neq k}\frac{1}{|C_l^v|}\sum\nolimits_{\hat{\boldsymbol h}^v_j \in C_l^v}  d(\hat{\boldsymbol h}^v_i,\hat{\boldsymbol h}^v_j),
\end{flalign}
where $C_k^v$ is the cluster where $h_i^v$ is located, $|C_k^v|$ is the number of instances in cluster $C_k^v$, 
and $d(\cdot,\cdot)$ refers to the distance between two instances. We select the view with the highest $SI$ as the best view $B$, calculated as:
\begin{equation}
SI(\hat{\boldsymbol H}^B) = \max\{SI(\hat{\boldsymbol H}^1),SI(\hat{\boldsymbol H}^2),...,SI({\hat{\boldsymbol H}^V})\}.\label{eq_bestview}
\end{equation}

\subsection{Dual-Weighted Contrastive Learning in the B-O Contrastive Mechanism} 
Considering view discrepancy, we propose the cross-view quality weight, denoted as $\mathcal{W}_{SI}^{v,B}$ and combine it with the view discrepancy weight~\cite{xu2024self} $\mathcal{W}_{CMI}^{v,B}$ to construct dual-weighted contrastive loss $\mathcal L_{CL}^{v,B}(\hat{\boldsymbol H}^v, \hat{\boldsymbol H}^B)$. 
The dual-weighted contrastive loss $\mathcal L_C^{v,B}(\hat{\boldsymbol H}^v, \hat{\boldsymbol H}^B)$ of the specific cross-view is expressed as:
\begin{equation}
    \mathcal L_C^{v,B}(\hat{\boldsymbol H}^v,\hat{\boldsymbol H}^B) = \mathcal{W}_{Dual}^{v,B}\cdot \mathcal L_{CL}^{(v,B)}(\hat{\boldsymbol H}^v, \hat{\boldsymbol H}^B),
\end{equation}
where the cross-view dual weight $\mathcal{W}_{Dual}^{v,B}=\mathcal{W}_{CMI}^{v,B}\cdot  \mathcal{W}_{SI}^{v,B}$. The loss $\mathcal L_{CL}^{v,B}$ take the form of previous contrastive loss InfoNCE~\cite{oord2018representation}.

In the dual-weighted contrastive loss, the cross-view quality weight is utilized to strengthen the high-quality cross-views and weaken the low-quality cross-views, defined as:
\begin{equation}
  \mathcal{W}_{SI}^{v,B} = \mathcal{W}_{SI}^v \cdot \mathcal{W}_{SI}^B,\label{eq_Wsi}
\end{equation}
where the view quality weight of the 
$v$-th view $\mathcal{W}_{SI}^v$ is expressed as: 
\begin{equation}
\mathcal{W}_{SI}^v=e^{SI(\hat{\boldsymbol H}^v)}, 
\end{equation}
Similarly, the view quality weight for the best view, denoted as $\mathcal{W}_{SI}^B$, is given by:
\begin{equation}
\mathcal{W}_{SI}^B=e^{SI(\hat{\boldsymbol H}^B)}. 
\end{equation}

\begin{algorithm}[tb]
\caption{Dual-Weighted Contrastive Learning.}
\label{alg:algorithm1}
\textbf{Input}: Dataset $\{\boldsymbol X_v\}_{v=1}^V$; Batch size $b$; Training epochs $T$; Iterations $I$; Parameters $\lambda$ and $\gamma$.\\
\textbf{Initialize:} the best view via Eq.~(\ref{eq_bestview}), $\{\mathcal{W}_{SI}^{v,B}\}_{v=1}^V$ and $\{\mathcal{W}_{CMI}^{v,B}\}_{v=1}^V$ like Eq.~(\ref{eq_Wsi}) and Eq.~(\ref{eq_cmi}) with $\{{\boldsymbol H}^v\}_{v=1}^V$, and $\{\theta^v,\eta^v\}_{v=1}^V$ by minimizing $\mathcal L_R$ in Eq.~(\ref{eq_rec}), respectively.\\
\begin{algorithmic}[1] 
\FOR{{$i=1$ to $I$}}
    \FOR{{$t=1$ to $T$}}{

        \STATE   Opitimize $\{\theta^v,\phi^v, \eta^v\}_{v=1}^V$ via Eq.~(\ref{eq_loss}) with Adam~\cite{fei2004learning};
	}
        
    \ENDFOR
    
    \STATE Compute cluster labels of each view by Eq.~(\ref{klabel});
    \STATE Update the best view by Eq.~(\ref{eq_bestview});
    \STATE Update $\{\mathcal{W}_{SI}^{v,B}\}_{v=1}^V$ with $\{\hat{\boldsymbol H}^v\}_{v=1}^V$ by Eq.~(\ref{eq_Wsi}) and $\{\mathcal{W}_{CMI}^{v,B}\}_{v=1}^V$ with $\{\hat{\boldsymbol H}^v\}_{v=1}^V$ by Eq.~(\ref{eq_cmi}) ;
        
\ENDFOR
\STATE Calculate clustering labels by executing k-means on $\{\hat{\boldsymbol h}_i\}_{i=1}^N$ obtained by Eq.~(\ref{feature}).
\end{algorithmic}
\textbf{Output}: The clustering labels $\hat{\boldsymbol Y} = [\hat{\boldsymbol y}_1, \hat{\boldsymbol y}_2,..., \hat{\boldsymbol y}_N]$; Encoder, decoder and MLP parameters $\{\theta^v, \phi^v, \eta^v\}_{v=1}^V$.
\end{algorithm}

Drawing upon SEM~\cite{xu2024self}, the view discrepancy weight $\mathcal{W}_{CMI}$ is used to strengthen the cross-views with low discrepancy and weaken the cross-views with high discrepancy. The discrepancy weight $\mathcal{W}_{CMI}$ of the best view and the other views is computed as:
\begin{equation}
\begin{aligned}
    & \mathcal{W}_{CMI}^{v,B} = e^{CMI(\hat{\boldsymbol y}^v,\hat{\boldsymbol y}^B)}-1,\\
   & CMI(\hat{\boldsymbol y}^v,\hat{\boldsymbol y}^B) = \frac{2I(\hat{\boldsymbol y}^v;\hat{\boldsymbol y}^B)}{H(\hat{\boldsymbol y}^v)+H(\hat{\boldsymbol y}^B)},\\
   & I(\hat{\boldsymbol y}^v;\hat{\boldsymbol y}^B) = H(\hat{\boldsymbol y}^v)-H(\hat{\boldsymbol y}^v|\hat{\boldsymbol y}^B),\label{eq_cmi}
\end{aligned}
\end{equation}
where $\hat{\boldsymbol y}^v\in R^N$ is the cluster label distribution of $K$ clusters in $v$-th view, $H(\hat{\boldsymbol y}^v) = -\sum_{k}^K P(\hat{\boldsymbol y}_k^v)log P(\hat{\boldsymbol y}_k^v)$ is the cross-entropy, $\hat{\boldsymbol y}_k^v \in \hat{\boldsymbol y}^v$, and $P(\hat{\boldsymbol y}_k^v)$is the marginal probability of $\hat{\boldsymbol y}_k^v$. 
However, the importance of cross-views composed of low-quality views with low discrepancy can be comparable to, or even higher than, those of cross-views composed of high-quality views with low discrepancy only according to view discrepancy weights. This results in the amplification of unreliable cross-views. 
Since the cross-view quality weight $\mathcal{W}_{SI}^{v,B}$ effectively mitigates the impact of low-quality cross-views, the dual weight $\mathcal{W}_{Dual}^{v,B}$ further diminishes the influence of cross-views that are low-discrepancy yet low-quality.

The overall loss function $\mathcal L$ is designed as:
\begin{equation}
    \mathcal L=\gamma \sum_{v=1}^V \mathcal L_C^{v,B} + \lambda \sum_{v=1}^V \mathcal L_R^{v},\label{eq_loss}
\end{equation}
where $\gamma$ and $\lambda$ are hyper-parameters, and the reconstruction loss $\mathcal L_R^{v}$ is expressed as:
\begin{equation}
   \mathcal L_R^{v} = \lambda \sum_{v=1}^V \Vert \boldsymbol{X}^v- \hat{\boldsymbol{X}}^v\Vert,\label{eq_rec}
\end{equation}
where $\Vert\cdot\Vert_2$ denotes vector $l_2$-norm, and $\{\hat{\boldsymbol{X}}^v \in R^{N \times d_v}\}_{v=1}^V$ is obtained by employing $v$-th decoder $f_D^v({\boldsymbol H}^v;\eta^v)$. Finally, the clustering label of the $i$-th instance $\hat{\boldsymbol y}_i$ is obtained by executing k-means on the feature of the $i$-th instance $\hat{\boldsymbol h}_i$ which is computed as:
\begin{equation}
\hat{\boldsymbol h}_i = \mathop{\Vert}\limits_{v=1}^{V}\, \hat{\boldsymbol h}_i^v,~\label{feature}
\end{equation}
where $\Vert$ represents concatenation operation.

Algorithm~\ref{alg:algorithm1} summarizes the training steps of DWCL. In the initialization stage, we first obtain low-level features $\{{\boldsymbol H}^v \in R^{N \times H^v}\}_{v=1 }^V$ by pre-training the model with Eq.(~\ref{eq_rec}). Then, initialize the weights $\{\mathcal{W}_{CMI}^{v,B}\}_{v=1}^V$ and $\{\mathcal{W}_{SI}^{v,B}\}_{v=1}^V$ with ${\boldsymbol H}^v$, and determine the best view by $\{\mathcal{W}_{SI}^{v,B}\}_{v=1}^V$. In the fine-tuning stage, the best view, along with the weights $\{\mathcal{W}_{CMI}^{v,B}\}_{v=1}^V$ and $\{\mathcal{W}_{SI}^{v,B}\}_{v=1}^V$ are alternately updated to promote each other.

\subsection{Theoretical Analysis}
For DWCL with InfoNCE loss and a dual weighting strategy, minimizing the contrastive loss effectively equals to maximize the mutual information among multiple views. Theorem 1 theoretically demonstrate that our contrastive loss offers a better optimization scope, as it ensures a superior lower bound for mutual information. Moreover, Theorem 2 demonstrates that the B-O contrastive mechanism can effectively weaken the impact of low-quality cross-views in contrastive losses and reduce computational complexity by decreasing the number of cross-views.

\textbf{Theorem 1} \textit{For any two views $v_i$, $v_j$, $\{v_i,v_j\}\in V$, letting the view quality weight $\mathcal W_{SI}^{v_i} = e^\alpha\in(1,e)$, and $\mathcal W_{SI}^{v_j} = e^\beta\in(1,e)$. The view discrepancy weight $\mathcal{W}_{CMI}^{v_i,v_j} \ge e^{\sigma/log N}-1$, where $\sigma$ donates the mutual information between high-level representation feature $\hat{\boldsymbol H}^{v_i}$ and $\hat{\boldsymbol H}^{v_j}$ and  $\sigma > 0$. Thus, minimizing the dual weighted contrastive loss $\mathcal{W}_{CMI}^{v_i,v_j}\cdot\mathcal W_{SI}^{v_i}\cdot\mathcal W_{SI}^{v_j}\cdot\mathcal L_{CL}^{(v_i,v_j)}(\hat{\boldsymbol H}^{v_i},\hat{\boldsymbol H}^{v_j})$ is equivalent to maximizing $e^{\alpha+\beta}(e^{\sigma/log N}-1)I(\hat{\boldsymbol H}^{v_i},\hat{\boldsymbol H}^{v_j})$, where $I(\hat{\boldsymbol H}^{v_i},\hat{\boldsymbol H}^{v_j})$ is the mutual information and $e^{\alpha+\beta} > 1$. Therefore, DWCL has better lower bounds than SEM.}

\textit{Proof} According to SEM~\cite{xu2024self}, the discrepancy weight $\mathcal{W}_{CMI}$ of cross-views is computed as Eq.~(\ref{eq_cmi}).
Let $\sigma > 0$ donate the mutual information $I(\hat{\boldsymbol y}^{v_i};\hat{\boldsymbol y}^{v_j})$, where $H(\hat{\boldsymbol y}^{v_i})+H(\hat{\boldsymbol y}^{v_j})$ is calculated as:
\begin{equation}
-\sum_{k\in N}P(\hat{\boldsymbol y}_k^{v_i})log P(\hat{\boldsymbol y}_k^{v_i})-\sum_{k\in N}P(\hat{\boldsymbol y}_k^{v_J})log P(\hat{\boldsymbol y}_k^{v_J})\le 2\log N.
\end{equation}
Thus, the discrepancy weight $\mathcal{W}_{CMI} \ge e^{\sigma/\log N}-1$. 

For the view quality weight $\mathcal W_{SI}^{v_i}$ of view ${v_i}$, it is expressed as:
\begin{equation}
\mathcal{W}_{SI}^{v_i}=e^{SI(\hat{\boldsymbol H}^{v_i})},
\end{equation}
where
\begin{flalign}
    &SI(\hat{\boldsymbol H}^{v_i})=\frac{1}{N}\sum_{\hat{\boldsymbol h}^{v_i}_i\in \hat{\boldsymbol H}^{v_i}}SI(\hat{\boldsymbol h}^{v_i}_i).
\end{flalign}
For a discussion of the range of $SI(\hat{\boldsymbol h}^{v_i}_i)$, if $\exists \,\hat{\boldsymbol h}^{v_i}_i \in \hat{\boldsymbol H}^{v_i}$ in cluster $C_k$, such that $a(\hat{\boldsymbol h}^{v_i}_i) > b(\hat{\boldsymbol h}^{v_i}_i)$, then the center of the cluster $C_k$ is closer to the points in the cluster $C_l$, contrary to the k-means~\cite{hartigan1979algorithm} algorithm. Thus, for $\forall\, \hat{\boldsymbol h}^{v_i}_i \in \hat{\boldsymbol H}^{v_i}$ in cluster $C_k$, $a(\hat{\boldsymbol h}^{v_i}_i) < b(\hat{\boldsymbol h}^{v_i}_i)$,
Let $\alpha$ and $\beta$ signify $SI(\hat{\boldsymbol H}^{v_i})$ and $SI(\hat{\boldsymbol H}^{v_j})$, where 
\begin{equation}
SI(\hat{\boldsymbol h}^{v_i}_i)\in (0,1), \quad i.e.\quad  SI(\hat{\boldsymbol H}^{v_i})\in (0,1). 
\end{equation}
The view quality weights $\mathcal W_{SI}^{v_i}$ and $\mathcal W_{SI}^{v_j}$ are expressed as:
\begin{equation}
\mathcal W_{SI}^{v_i}=e^\alpha\in(1,e),\quad
\mathcal W_{SI}^{v_j}=e^\beta\in(1,e).
\end{equation}
As proven in~\cite{oord2018representation}, minimizing the contrastive losses $\mathcal L_{CL}^{(v_i,v_j)}(\hat{\boldsymbol H}^{v_i},\hat{\boldsymbol H}^{v_j})$ equals to maximizing the mutual information $I(\hat{\boldsymbol H}^{v_i},\hat{\boldsymbol H}^{v_j})$. 
Thus, minimizing the dual weighted contrastive loss $\mathcal L_C^{v_i,v_j}(\hat{\boldsymbol H}^{v_i}, \hat{\boldsymbol H}^{v_j})$, computed as:
\begin{equation}
   \mathcal L_C^{v_i,v_j}(\hat{\boldsymbol H}^{v_i}, \hat{\boldsymbol H}^{v_j}) = \mathcal{W}_{CMI}^{v_i,v_j}\cdot\mathcal W_{SI}^{v_i}\cdot\mathcal W_{SI}^{v_j}\cdot\mathcal L_{CL}^{(v_i,v_j)}(\hat{\boldsymbol H}^{v_i},\hat{\boldsymbol H}^{v_j}),
\end{equation}
 is equivalent to maximizing:
 \begin{equation}
\mathcal L_C^{v_i,v_j}(\hat{\boldsymbol H}^{v_i}, \hat{\boldsymbol H}^{v_j}) =  e^{\alpha+\beta}(e^{\sigma/log N}-1)I(\hat{\boldsymbol H}^{v_i},\hat{\boldsymbol H}^{v_j}),
 \end{equation}
where $e^{\alpha+\beta} > 1$, which completes the proof.

\textbf{Theorem 2} \textit{In the traditional pairwise contrastive mechanism, cross-view quality weight $\mathcal{W}_{SI}^{v_i,v_j}$ is computed as $\sum_{v_j\in V}\mathcal W_{SI}^{v_i}\cdot\mathcal W_{SI}^{v_j}\in (V, Ve^2)$. In the B-O contrastive mechanism, cross-view quality weight $\mathcal{W}_{SI}^{v,B}$ is computed as $\mathcal W_{SI}^{v}\cdot\mathcal W_{SI}^{B}\in (1, e^2)$, which utilizes a more efficient contrastive mechanism compared to the traditional pairwise contrastive approach, preventing low-quality cross-views from being augmented. The B-O contrastive mechanism effectively safeguards the representation quality of multi-views by mitigating the impact of low-quality views.}

\textit{Proof} 
In the traditional pairwise contrastive mechanism, the total contrastive loss $\mathcal L_{C}^{(v_i,v_j)}(\hat{\boldsymbol H}^{v_i},\hat{\boldsymbol H}^{v_j})$ is expressed as:
\begin{equation}
    \sum_{v_i\in V}\sum_{v_j\in V}\mathcal{W}_{CMI}^{v,B}\cdot \mathcal{W}_{SI}^{v_i,v_j}\cdot\mathcal L_{CL}^{(v_i,v_j)}(\hat{\boldsymbol H}^{v_i},\hat{\boldsymbol H}^{v_j}),
\end{equation}
where the quantity of cross-views is $V^2$.
For view $v_i\in V$, cross-view quality weight $\mathcal{W}_{SI}^{v_i,v_j}$ is computed as:
\begin{equation}
\mathcal{W}_{SI}^{v_i,v_j} = \sum_{v_j\in V}\mathcal W_{SI}^{v_i}\cdot\mathcal W_{SI}^{v_j}\in (V, Ve^2). 
\end{equation}
In the B-O contrastive mechanism, if $SI(\hat{\boldsymbol H}^{v_i})$ or $SI(\hat{\boldsymbol H}^{v_j})$ is not maximum in $\{SI(\hat{\boldsymbol H}^1),SI(\hat{\boldsymbol H}^2),...,SI({\hat{\boldsymbol H}^V})\}$, letting $\mathcal{W}_{SI}^{v_i,v_j}=0$, \textit{i.e.} the total contrastive loss $\mathcal L_{C}^{(v,B)}(\hat{\boldsymbol H}^{v},\hat{\boldsymbol H}^{B})$ is calculated as:
\begin{equation}
\mathcal L_{C}^{(v,B)}(\hat{\boldsymbol H}^{v},\hat{\boldsymbol H}^{B}) = \sum_{v\in V}\mathcal{W}_{CMI}^{v,B}\cdot\mathcal{W}_{SI}^{v,B}\cdot\mathcal L_{CL}^{(v,B)}(\hat{\boldsymbol H}^{v},\hat{\boldsymbol H}^{B}),
\end{equation}
where the quantity of cross-views is just $V$, the complexity is reduced from quadratic to linear, $i.e.$, from $O(|V|^2) $ to $O(|V|)$, decreasing the computational costs. 
For view $v\in V$, cross-view quality weight $\mathcal{W}_{SI}^{v,B}$ is computed as:
\begin{equation}
\mathcal{W}_{SI}^{v,B} = \mathcal W_{SI}^{v}\cdot\mathcal W_{SI}^{B}\in (1, e^2). 
\end{equation}
Therefore, we employ a more efficient contrastive mechanism compared to the traditional pairwise contrastive approach, effectively safeguarding the representation quality of multi-views by mitigating the impact of low-quality views.

\subsection{Complexity Analysis} 
Let $b$, $V$, $I$, $d_v$, and $\hat{h}_v$ donate the batch size, number of views, iterations, original feature dimension, and the higher hierarchical feature dimension respectively. The total training epochs are $E = I \times T$, where $T$ is the training epochs of each iteration. For each iteration, the view discrepancy weights $\{\mathcal W_{CMI}^{v,B}\}_{v=1}^{V}$ are updated. For each view $v$, the time complexity of k-means is $O(IbK\hat{h}_v)$, so the total time complexity of this step is $O(VIbK\hat{h}_v) $. The time complexity of updating the cross-view quality weights $\{\mathcal W_{SI}^{v,B}\}_{v=1}^{V}$ is $O(Vb^2\hat{h}_v)$. For each epoch, the time complexity of the reconstruction loss $\mathcal L_R$ is $O(Vb\,d_v)$. In the B-O contrastive mechanism, the time complexity of the contrastive loss $\mathcal L_c$ is $O(Vb^2\hat{h}_v)$, which is different from the time complexity $O(V^2b^2\hat{h}_v)$ in the traditional pairwise contrastive mechanism. The complexity is reduced from $O(|V|^2) $ to $O(|V|)$, decreasing the computational costs. 
Therefore, the overall time complexity of DWCL is $I(VbK\hat{h}_v + Vb^2\hat{h}_v) + E(Vb\,d_v + Vb^2\hat{h}_v)$.

\begin{table}
\setlength\tabcolsep{8pt}
    \centering
        \caption{
        Details of the Eight Multi-View Datasets.}
    \begin{tabular}{l|c|c|c}
    \hline
        \textbf{Datasets} & \textbf{Instances} & \textbf{Views} & \textbf{Classes} \\
    \hline
        Caltech5V7 & 1,400 & 5 & 7  \\
        Caltech6V7 & 1,400 & 6 & 7  \\
        Caltech6V20 & 2,386 & 6 & 20  \\
        DHA & 483 & 2 & 23  \\
        NUSWIDE& 5,000 & 5 & 5  \\
        Scene & 4,485 & 3 &  15 \\
        Fashion & 10,000 & 3 & 10  \\
        CIFAR10 & 60,000 & 7 &10 \\
    \hline
    \end{tabular}
    \Description{}
    \label{tab:datasets}
\end{table}

\begin{table}

    \centering
    \Description{Description of Descriptors Across All Datasets.}
\caption{
Description of Descriptors Across All Datasets.}
    
    \begin{tabular}{l|c}
        
    \hline
    \textbf{Descriptor}     & \textbf{Description}\\
    \hline
     Wavelet Moments     &Based on wavelet transform \\
     CENTRIST    &Characteristics of local texture structure \\
    HOG     & Histogram of Oriented Gradient\\
    GIST     &Description of macro level scene features \\
    LBP     &Local Binary Patterns \\
    Gabor  & Linear filter for edge extraction \\
    CH  &   Color Histogra \\ 
    BWCM  &  Block-Wise Color Moments \\   
   EDH   & Edge Direction Histogram \\      
   WT   &  Wavelet Texture  \\ 
   CMT   & A motion tracking algorithm \\      
   SIFT   &  Scale-Invariant Feature Transform\\      
   
    \hline
    \end{tabular}
    \label{tab:descriptor}
\end{table}

\section{Experiments}

\subsection{Experimental Settings}

\textbf{Datasets} 
Our experiments utilize eight multi-view datasets, as shown in Tab.~\ref{tab:datasets}. 
Specifically, we use the Caltech dataset~\cite{fei2004learning}, focusing on Caltech5V7, which contains 1,400 instances represented by five views: Wavelet Moments, CENTRIST, HOG, GIST, and LBP across seven categories. 
Caltech6V7 and Caltech6V20 expand on this with six views—Gabor, Wavelet Moments, CENTRIST, HOG, GIST, and LBP—featuring 1,400 instances across seven categories and 2,386 instances across 20 categories, respectively. 
The DHA dataset~\cite{lin2012human}, focusing on human action, comprises depth and RGB views with 483 instances across 23 categories. Similarly, NUSWIDE~\cite{jiang2011consumer} includes 5,000 instances across five categories, represented by five views: CH, BWCM, Colour Correlogram, EDH, and WT. The Scene dataset~\cite{fei2005bayesian} includes 4,485 images representing 15 scene categories, each depicted through three views. The Fashion dataset~\cite{xiao2017fashion} comprises 10,000 images across 10 categories of commodities, each with three views. The CIFAR10 dataset~\cite{krizhevsky2009learning} consists of 60,000 images spanning 10 categories of objects, with each category featuring images captured from seven different perspectives. 
Additionally, Tab.~\ref{tab:descriptor} provides a brief overview of the descriptors used for these views.

\begin{table}
 \renewcommand\arraystretch{1.2}
\setlength\tabcolsep{1pt}

    \centering
 \caption{
 Network Hyper-parameters for All Datasets.}
 \scalebox{0.7}{
    \begin{tabular}{l|c|c|c}

    \hline
    \textbf{Dataset}     & \textbf{Dimensions of Input and Output}  & \textbf{Dimensions of} $H$ &\textbf{ Dimensions of} $\hat{H}$\\
    \hline
     Caltech5V7    & [40, 254, 1984, 512, 928] & 512 & 128 \\
     Caltech6V7    & [48, 40, 254, 1984, 512, 928] & 512 & 128\\
     Caltech6V20    & [48, 40, 254, 1984, 512, 928] & 512 &  128 \\
     DHA    & [110, 6144] & 512 & 512\\
     NUSWIDE    & [64, 225, 144, 73, 128] & 512 & 96\\
     Scene    &[20, 59, 40]  & 512 &128 \\
     Fashion   &[784, 784, 784]  & 512 & 128 \\
     CIFAR10 & [3072, 5376, 512, 5376, 5376, 1239, 5376] & 512 &128\\
    \hline
    \end{tabular}}
    \label{tab:network hyper-parameters}
\end{table}

\begin{table}
\setlength\tabcolsep{2pt}
    \centering
 \caption{
 Training Hyperparameters for All Datasets.}
           
   {\begin{tabular}{l|c|c|c}

    \hline
    \textbf{Dataset}     & \textbf{Pre-training Epochs} & \textbf{CL Iterations} & \textbf{CL Epochs} \\
    \hline
     Caltech5V7    & 100 & 3 &50  \\
     Caltech6V7    & 100 & 3 & 50\\
     Caltech6V20    & 100 & 4 & 25  \\
     DHA    & 100 & 1 & 50\\
     NUSWIDE    & 100 & 5 & 20\\
     Scene    & 100 & 4 & 100 \\
     Fashion   & 100 & 6 & 50 \\
     CIFAR10 & 50 & 3 & 10 \\
    \hline
    \end{tabular}}
    \label{tab:training hyperparameters}
\end{table}

\begin{figure}
    \centering
    \includegraphics[width=1\linewidth]{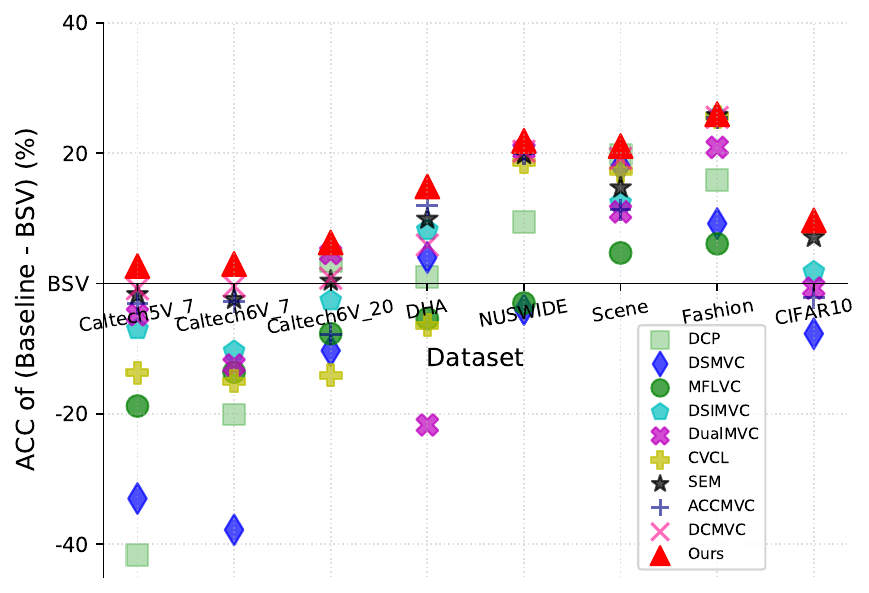}
    \Description{}
    \caption{
    Comparison of DWCL with seven baseline methods and BSV across eight datasets. The vertical axis represents the difference in ACC between each baseline and the BSV for eight multi-view datasets.
    }
    \label{fig:scatter}
\end{figure}

\begin{table*}
 \renewcommand\arraystretch{1.15}
    \centering
     \caption{
     Clustering performance comparisons across eight multi-view datasets (mean$\pm$std\%). The BSV represents k-means clustering performance based on the best single view. The top results are highlighted in bold.
     } 
    \begin{tabular}{l|cccccccccccc}
    \hline
    \multirow{2.5}{*}{\textbf{Method}}
    & \multicolumn{2}{c}{\textbf{Caltech5V7}}& \multicolumn{2}{c}{\textbf{Caltech6V7}}& \multicolumn{2}{c}{\textbf{Caltech6V20}}& \multicolumn{2}{c}{\textbf{DHA}}
    \\
\cline{2-9}
       & \textbf{ACC} & \textbf{NMI}  & \textbf{ACC} & \textbf{NMI}  & \textbf{ACC} & \textbf{NMI} & \textbf{ACC} & \textbf{NMI} \\
     \hline
     BSV & {}{89.9$\pm$0.9} & {}{81.9$\pm$1.5} & {}{89.3$\pm$1.2} & {}{81.1$\pm$1.7} &46.1$\pm$1.3 & 60.6$\pm$0.5 &  68.4$\pm$2.6 &  79.4$\pm$0.8  \\
     DCP~\cite{lin2022dual} & 48.2$\pm$2.4 & 51.9$\pm$1.7 & 69.8$\pm$1.1 & 65.5$\pm$2.5 & 49.6$\pm$3.7 & 50.7$\pm$2.8 & 69.4$\pm$2.3&  82.6$\pm$2.9  \\
DSMVC~\cite{tang2022deep}& 56.9$\pm$2.7 & 49.1$\pm$2.1 &  51.5$\pm$4.4 & 41.5$\pm$6.3 &35.8$\pm$5.1 & 45.3$\pm$1.5 & 72.4$\pm$5.6 &  79.2$\pm$1.3  \\
DSIMVC~\cite{tang2022DSI} & 71.1$\pm$1.5 & 59.9$\pm$1.5 & 75.7$\pm$1.2 & 63.8$\pm$1.4 & 38.4$\pm$1.1 & 45.5$\pm$2.0 & 63.0$\pm$2.7 &  74.7$\pm$3.2  \\
MFLVC~\cite{xu2022multi} & 82.9$\pm$0.2 & 75.2$\pm$1.3 & 78.9$\pm$4.9 & 69.6$\pm$5.4 &43.5$\pm$2.9 & 58.4$\pm$2.0 &  76.6$\pm$3.8 &  82.4$\pm$1.4  \\
DualMVC~\cite{yang2023dealmvc} & 85.1$\pm$0.8 & 76.1$\pm$1.0 & 76.8$\pm$0.3 & 67.2$\pm$0.9 & {}{50.7$\pm$0.3} & {57.3$\pm$0.6} & 46.7$\pm$5.3 & 69.6$\pm$7.5  \\
CVCL~\cite{chen2023deep} & 76.2$\pm$2.3 & 66.4$\pm$1.2 & 74.3$\pm$4.8 & 66.4$\pm$3.7 & 32.0$\pm$5.4 & 51.5$\pm$1.8 & 62.0$\pm$2.2 & 64.2$\pm$3.0 \\	
SEM~\cite{xu2024self} & 88.2$\pm$1.4 & 80.3$\pm$1.6 & 86.9$\pm$0.4 & 79.2$\pm$1.4 & 46.5$\pm$1.9 & {}{62.5$\pm$0.3} & {}{78.3$\pm$2.9} &  {}{83.1$\pm$0.5}  \\
ACCMVC~\cite{yan2024anchor} & 86.8$\pm$1.9&	80.4$\pm$2.6		&	86.5$\pm$1.5&	76.7$\pm$3.2	&		38.3$\pm$1.2	&58.5	$\pm$2.5&		80.4$\pm$1.5	&81.0 $\pm$0.9\\
DCMVC~\cite{cui2024dual} & 89.1$\pm$1.7&	81.7$\pm$2.3	&		88.9$\pm$3.1	&81.3$\pm$2.6	&		46.9$\pm$1.8	&63.3$\pm$0.9	&		74.3$\pm$3.3	&80.2$\pm$2.1\\
 \textbf{DWCL (Ours)} &\textbf{92.6$\pm$1.5} & \textbf{85.5$\pm$3.3} & \textbf{92.3$\pm$2.6} & \textbf{86.2$\pm$2.1} & \textbf{52.4$\pm$1.7} & \textbf{63.7$\pm$1.8} & \textbf{83.3$\pm$1.9} & \textbf{85.0$\pm$0.8} 
\\
\hline
\\
    \hline
    \multirow{2.5}{*}{\textbf{Method}}
& \multicolumn{2}{c}{\textbf{NUSWIDE}}&\multicolumn{2}{c}{\textbf{Fashion}}& \multicolumn{2}{c}{\textbf{Scene}}   & \multicolumn{2}{c}{\textbf{CIFAR10}}
    \\
\cline{2-9}

       & \textbf{ACC} & \textbf{NMI}  & \textbf{ACC} & \textbf{NMI}  & \textbf{ACC} & \textbf{NMI} & \textbf{ACC} & \textbf{NMI} \\
     \hline
     BSV & 40.3$\pm$1.3 & 17.4$\pm$0.8 & 69.9$\pm$2.1 & 60.4$\pm$2.7 & 22.4$\pm$0.8 & 23.4$\pm$1.8 & 29.3$\pm$2.9 &  16.9$\pm$1.4  \\
     DCP~\cite{lin2022dual} & 49.7$\pm$1.8 & 24.7$\pm$3.1 & 89.6$\pm$0.8 &  89.1$\pm$0.7 & {}{42.2$\pm$1.0} & 39.9$\pm$1.3  & - & - \\
DSMVC~\cite{tang2022deep}& 36.3$\pm$2.5 & {}{36.4$\pm$1.9}& 82.9$\pm$6.5 &  80.1$\pm$4.9  & 40.4$\pm$6.1 & {}{42.1$\pm$4.1}  & 21.6$\pm$2.0 & 10.6$\pm$3.4 \\
DSIMVC~\cite{tang2022DSI} & 37.3$\pm$3.2 & 12.0$\pm$2.1 & 79.8$\pm$3.1 &  77.8$\pm$2.3 & 27.1$\pm$0.6 & 28.0$\pm$0.5 & - & -  \\
MFLVC~\cite{xu2022multi} & 61.0$\pm$2.7 & 34.7$\pm$2.4 & 99.3$\pm$0.0 &  98.2$\pm$0.1 & 34.8$\pm$1.8 & 36.7$\pm$1.3  & 31.1$\pm$2.3 & {}{24.9$\pm$2.1} \\
DualMVC~\cite{yang2023dealmvc} & {}{60.7$\pm$5.2} & {33.9$\pm$6.8}& 94.6$\pm$0.1 &  87.5$\pm$0.1  & 33.4$\pm$5.1& 34.4$\pm$3.5 & 28.6$\pm$2.8 & 20.2$\pm$3.3  \\
CVCL~\cite{chen2023deep} & 58.9$\pm$4.1 & 30.6$\pm$3.0 & {99.3$\pm$0.0} &  {}{98.3$\pm$0.0} & 39.7$\pm$1.7 & 41.0$\pm$3.3  & -& - \\	
SEM~\cite{xu2024self} & 60.1$\pm$1.7 & 33.6$\pm$1.7 & {}{99.4$\pm$0.0} &  98.2$\pm$0.0 & 37.1$\pm$2.9 & 39.2$\pm$2.7 & {}{34.6$\pm$1.9} & 24.6$\pm$2.4  \\
ACCMVC~\cite{yan2024anchor} &  60.2$\pm$3.4&	35.7	$\pm$1.7	&	99.2$\pm$0.6&98.0$\pm$1.0	&	33.7$\pm$3.2	&35.5$\pm$1.8	&		27.1$\pm$2.2	&17.0$\pm$1.5
 \\
DCMVC~\cite{cui2024dual} & 60.6$\pm$2.7&	30.3$\pm$0.6	&		99.3$\pm$0.2	&98.2$\pm$0.8	&		41.6$\pm$3.6 &	43.5$\pm$1.4	&		-	&-
  \\
 \textbf{DWCL (Ours)} &\textbf{62.2$\pm$1.3} & \textbf{37.2$\pm$1.3} & \textbf{99.5$\pm$0.1} & \textbf{98.6$\pm$0.3}  & \textbf{43.5$\pm$2.1} & \textbf{44.7$\pm$1.9} & \textbf{39.0$\pm$1.2} & \textbf{25.6$\pm$1.5}
 
\\
\hline

    \end{tabular}
    \label{tab:comparison2}
\end{table*}

\textbf{Evaluation Metrics} 
To evaluate the effectiveness of our DWCL, we use two widely adopted metrics: clustering accuracy (ACC) and normalized mutual information (NMI). ACC measures the proportion of data points assigned to the correct cluster by optimally matching predicted clusters to true labels, indicating how well the clustering algorithm aligns with the true class labels. NMI, a more robust metric, assesses the shared information between true labels and predicted cluster assignments, normalizing the mutual information score to range from 0 (no mutual information) to 1 (perfect agreement). This normalization accounts for variations in clustering results due to the number or distribution of clusters. Higher values of ACC and NMI signify better clustering performance.

\textbf{Implementation Details} 
All experiments are conducted on an NVIDIA GeForce RTX 3090 GPU (24 GB) and an Intel(R) Xeon(R) Gold 6230R CPU @ 2.10 GHz. We employ an encoder-decoder network structure, with both the encoder and decoder consisting of a four-layer fully connected network, following previous works~\cite{xu2022multi,tang2022deep,yang2023dealmvc}.
The encoded feature ${H}$ is passed through an MLP layer to produce $\hat{H}$, which is then fed to the decoder to obtain $\hat{X}$.
The network encoding process is expressed as $X^i$ $\rightarrow$ 500 $\rightarrow$ ReLU $\rightarrow$ 500 $\rightarrow$ ReLU $\rightarrow$ 2000 $\rightarrow$ $H$ $\rightarrow$ $\hat{H}$, while the decoding process is $H$ $\rightarrow$ 200 $\rightarrow$ ReLU $\rightarrow$ 500 $\rightarrow$ ReLU $\rightarrow$ 500 $\rightarrow$ $\hat{X}$.
Here, ReLU~\cite{glorot2011deep} serves as the activation function, and $X^i$ represents the $i$-th view.
Tab.~\ref{tab:network hyper-parameters} provides the network hyperparameters for each dataset. The training process consists of a pre-training stage followed by fine-tuning, with specific parameters detailed in Tab.~\ref{tab:training hyperparameters}. Reconstruction loss is used during pre-training to determine initial view quality weights and identify the best view based on the silhouette coefficient. In the fine-tuning stage, we incorporate a dual-weighted InforNCE contrastive loss. Hyperparameters $\gamma$ and $\lambda$ are set to 1.0, with Adam~\cite{kingma2014adam} as the optimizer, a learning rate of 0.0003, and a batch size of 128 across all datasets.

\begin{figure*}[h]
\centering 
\subfloat[Initial View ACC]{
\label{fig:view_acc}
\includegraphics[width=0.32\textwidth]{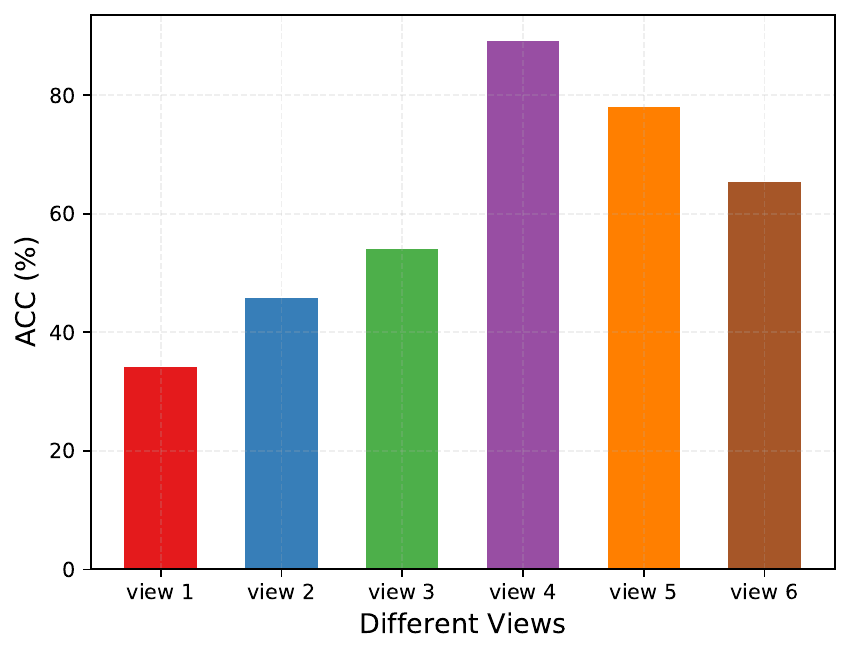}}
\subfloat[Loss Weight $vs.$ Iteration]{
\label{fig:Weight_loss}
\includegraphics[width=0.32\textwidth]{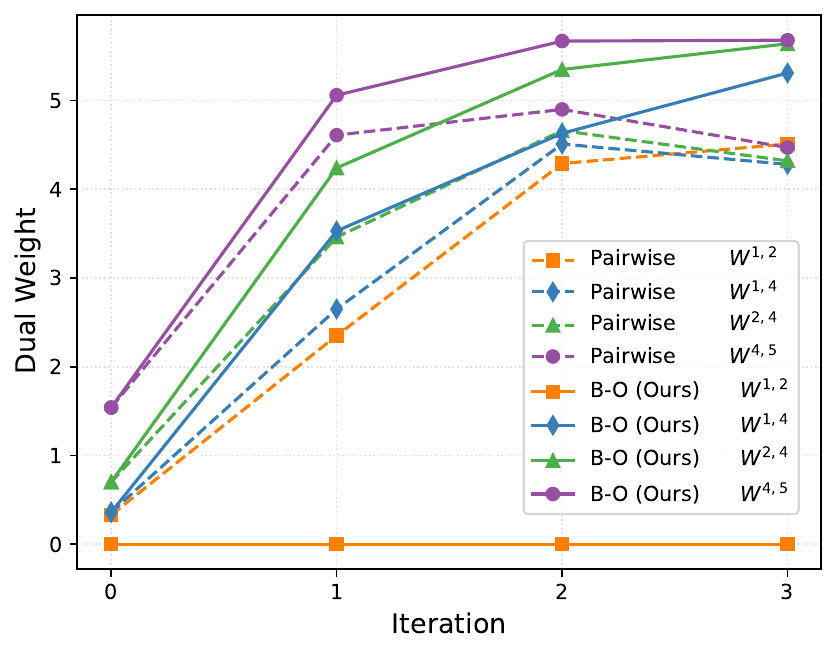}}
\subfloat[ACC $vs.$ Iteration]{
\label{fig:Each_view}
\includegraphics[width=0.32\textwidth]{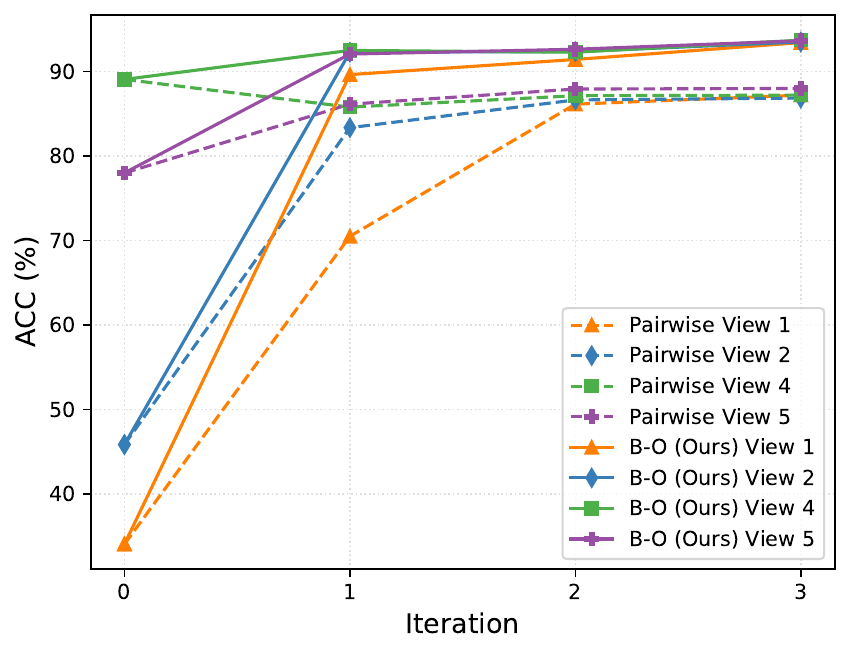}}
\caption{
(a) Initial accuracy of each view on Caltech6V7.
(b) Changes in weights $\mathcal{W}^{1,2}$, $\mathcal{W}^{1,4}$, $\mathcal{W}^{2,4}$ and $\mathcal{W}^{4,5}$ in the traditional pairwise and B-O contrastive mechanisms.
(c) Clustering performance (ACC) of views 1, 2, 4, and 5 in the traditional pairwise and B-O contrastive mechanisms.
} 
    \Description{}

\label{fig:weight}
\end{figure*}

\begin{table*}
 \renewcommand\arraystretch{1.15}
\setlength\tabcolsep{4pt}
    \centering
            \caption{
            Clustering performance of various cross-view contrastive mechanisms across eight multi-view datasets.} 
    \begin{tabular}{l|cccccccccccccccc}
    \hline
    \multirow{2.5}{*}{\textbf{Method}}& \multicolumn{2}{c}{\textbf{Caltech5V7}}
& \multicolumn{2}{c}{\textbf{Caltech6V7}}& \multicolumn{2}{c}{\textbf{Caltech6V20}}& \multicolumn{2}{c}{\textbf{DHA}}& \multicolumn{2}{c}{\textbf{NUSWIDE}}& \multicolumn{2}{c}{\textbf{CIFAR10}}& \multicolumn{2}{c}{\textbf{Scene}}& \multicolumn{2}{c}{\textbf{Fashion}}
    \\
\cline{2-17}
    
           &\textbf{ACC}&\textbf{NMI} & \textbf{ACC} & \textbf{NMI} & \textbf{ACC} & \textbf{NMI}& \textbf{ACC} & \textbf{NMI}  & \textbf{ACC} & \textbf{NMI}  & \textbf{ACC} & \textbf{NMI}  &\textbf{ACC}&\textbf{NMI} &\textbf{ACC}&\textbf{NMI}  \\
     \hline
     Pairwise&91.3 & 82.8&  88.0 & 81.2 &  50.2 & 61.6 &  75.4 & 82.2& 59.2& 29.7 & 38.1	&24.1 & 40.2 & 40.6 & 99.3&98.2 \\
     \textbf{B-O (Ours)} & \textbf{92.6}&\textbf{85.5} & \textbf{92.3} & \textbf{86.2} & \textbf{52.4} &\textbf{63.7} & \textbf{83.3}& \textbf{85.0} & \textbf{62.2}&\textbf{37.2} & \textbf{39.0}& \textbf{24.6} &\textbf{43.4} & \textbf{44.7} & \textbf{99.5}& \textbf{98.6}
 \\
\hline
    \end{tabular}
    \label{tab:best_view}
\end{table*}

\begin{table*}
 \renewcommand\arraystretch{1.2}
\setlength\tabcolsep{2pt}
     \caption{
     Comparison of clustering accuracy for each view among our DWCL, CVCL, and SEM methods on the Caltech5V7 and NUSWIDE datasets. The Silhouette Coefficient for each view is calculated after pre-training and multiplied by a factor of 10 for easier comparison. \textit{Initialization (ACC)} refers to each view's clustering performance during the pre-training stage. \textit{Relative Improvement} indicates the percentage increase in ACC for each view after fine-tuning, relative to the initialization. A red up arrow signifies improvement, while a blue down arrow indicates a decrease in ACC.
     } 
    \begin{tabular}{llccccccccccccccc}
    \hline
    \multicolumn{2}{c}{\textbf{Method}} 
& \multicolumn{5}{c}{\textbf{Caltech5V7}} & \multicolumn{5}{c}{\textbf{NUSWIDE}}
    \\

      &   &\textbf{View 1} & \textbf{View 2} & \textbf{View 3} & \textbf{View 4} & \textbf{View 5}&\textbf{View 1} & \textbf{View 2} & \textbf{View 3}& \textbf{View 4} & \textbf{View 5} \\
     \hline

\multicolumn{2}{c}{\textbf{Silhouette Coefficient }($\times$10)} & 2.2 & 1.7 & 3.0 & 1.8 & 2.6 & 0.8 & 1.0 & 0.6 & 0.6 & 1.7\\
     \hline
     
\multicolumn{2}{c}{\textbf{Initialization (ACC)}} & 45.7 & 56.9 & 89.9 & 75.2 & 65.2 & 32.4 & 30.0 & 36.1 & 38.7 & 39.1\\
     
     \hline

{\multirow{2}{*}{CVCL}}  
 & Fine-tuning (ACC) & 71.1 & 	73.8 & 	76.8	 & 76.5	 & 76.9	 & 52.1	 & 54.8	 & \textbf{59.7}	 & 58.7	 & 59.3
  \\
 &Relative Improvement (\%) &  55.5 ({\textcolor{red}{$\uparrow$}}) 	& 29.6 (\textcolor{red}{$\uparrow$}) & 	14.6 (\textcolor{blue}{$\downarrow$}) & 	1.7  (\textcolor{red}{$\uparrow$})	& 17.9 (\textcolor{red}{$\uparrow$}) 	& 60.8 (\textcolor{red}{$\uparrow$}) 	& 82.4 (\textcolor{red}{$\uparrow$}) 	& \textbf{65.5} (\textcolor{red}{$\uparrow$}) 	& 51.7 (\textcolor{red}{$\uparrow$}) &	51.6 (\textcolor{red}{$\uparrow$}) 
 \\
 
     \hline
{\multirow{2}{*}{SEM}}  
 & Fine-tuning (ACC) &  90.1 	& 	84.6 	& 	87.7 	& 	87.7 	& 	88.7 	& 	56.8 	& 	56.4 	& 	54.3 	& 	54.0 	& 	54.6 
 \\
 &Relative Improvement (\%) &   97.1 (\textcolor{red}{$\uparrow$}) & 	48.7 (\textcolor{red}{$\uparrow$}) 	& 2.5 (\textcolor{blue}{$\downarrow$}) 	& 16.6 (\textcolor{red}{$\uparrow$}) 	& 36.0 (\textcolor{red}{$\uparrow$}) & 	75.3 (\textcolor{red}{$\uparrow$}) 	& 87.8 (\textcolor{red}{$\uparrow$}) 	& 50.5 (\textcolor{red}{$\uparrow$}) & 	39.6  (\textcolor{red}{$\uparrow$})	& 39.6  (\textcolor{red}{$\uparrow$})
\\
 
     \hline
{\multirow{2}{*}{\textbf{DWCL (Ours)}}}  
 & Fine-tuning (ACC) &   \textbf{93.4} 	& \textbf{93.2} 	& \textbf{93.4} 	& \textbf{93.1} 	& \textbf{93.2 }	& \textbf{59.1} 	& \textbf{59.2} 	& 59.5 	& \textbf{59.9} 	& \textbf{60.4} 
\\
 &Relative Improvement (\%) &   \textbf{104.2} (\textcolor{red}{$\uparrow$}) 	& \textbf{63.7} (\textcolor{red}{$\uparrow$}) 	& \textbf{3.8} (\textcolor{red}{$\uparrow$}) 	& \textbf{23.8} (\textcolor{red}{$\uparrow$}) 	& \textbf{42.9} (\textcolor{red}{$\uparrow$}) 	& \textbf{82.4} (\textcolor{red}{$\uparrow$}) 	& \textbf{97.1} (\textcolor{red}{$\uparrow$}) 	& 65.0 (\textcolor{red}{$\uparrow$}) 	& \textbf{54.7} (\textcolor{red}{$\uparrow$}) 	& \textbf{54.4} (\textcolor{red}{$\uparrow$})
\\
 
\hline

    \end{tabular}
    \label{tab:view_improvement}
    \vspace{0.05cm}
\end{table*}

\subsection{Performance Comparison and Analysis}%
We compare our method with BSV and nine state-of-the-art deep Multi-View Contrastive Clustering (MVCC) methods. The BSV is trained solely with reconstruction loss, generating clustering results by applying k-means on the best view. The nine evaluated deep MVCC methods are DCP~\cite{lin2022dual}, DSMVC~\cite{tang2022deep}, DSIMVC~\cite{tang2022deep}, MFLVC~\cite{xu2022multi}, DualMVC~\cite{yang2023dealmvc}, CVCL~\cite{chen2023deep}, SEM~\cite{xu2024self}, ACCMVC~\cite{yan2024anchor}, and DCMVC~\cite{cui2024dual}.

The clustering performance on eight multi-view datasets is presented in Tab.~\ref{tab:comparison2}, averaged over 10 runs. DWCL outperforms all compared methods, demonstrating its superiority. Specifically, our method achieves an accuracy that exceeds the second-best method by approximately 3.5\%, 3.4\%, 2.9\%, and 4.4\% on Caltech5V7, Caltech6V7, DHA, and CIFAR10, respectively.
The clustering performance of the BSV method reflects the representation capability of the best single view without using contrastive learning. This indicates representation degeneration if the clustering performance of MVCC methods is inferior to that of the BSV.

\begin{table*}
\setlength\tabcolsep{8pt}
    \centering
     \caption{
     Clustering performance of DWCL with various weights and cross-View contrastive learning methods across eight multi-view datasets.
     } 
       
    \begin{tabular}{llcccccccc}
    \hline
    \multicolumn{2}{c}{\multirow{2}{*}{\textbf{Method}}}
& \multicolumn{2}{c}{\textbf{Caltech5V7}} & \multicolumn{2}{c}{\textbf{DHA}}& \multicolumn{2}{c}{\textbf{Scene}} 
 & \multicolumn{2}{c}{\textbf{Caltech6V7}}    \\

      &   &\textbf{ACC} &\textbf{NMI} & \textbf{ACC} & \textbf{NMI} &  \textbf{ACC} & \textbf{NMI}  &  \textbf{ACC} & \textbf{NMI} \\
     \hline
{\multirow{4}{*}{Pairwise}} & w/o $\mathcal{W}$ &  85.5&	76.9&			75.6&	82.8	&		39.4&	41.1&85.6 &	76.7 	\\
& w/ $\mathcal{W_{CMI}}$  & 89.3&	84.0	&		78.8&	{}{84.5}	&	36.7&	38.9 & 87.3 &	80.0 \\
& w/ $\mathcal{W_{SIL}}$ & 90.1&	83.5	&	80.5&	83.9&	38.7&	40.1 & 85.6 &	77.0 \\
& w/ $\mathcal{W_{CMI}}$+$\mathcal{W_{SIL}}$  &90.8&	83.3	&	78.5&	83.8&			40.2&	40.6& 87.6 	& 80.5 \\

      \hline

{\multirow{4}{*}{\textbf{B-O (Ours)}}} & w/o $\mathcal{W}$ &  89.3&	81.3	&77.6	&81.2	&	40.8&	42.9 & 90.9 &	85.8 \\
& w/ $\mathcal{W_{CMI}}$  & 91.3&	83.0	&	{82.0}	&82.9&41.3	&42.8 & 92.2 &	86.3 \\
& w/ $\mathcal{W_{SIL}}$  &  {}{91.8}&	{}{83.9}&		78.5&	83.0	&		{}{41.9}&	{}{43.5}& 92.4 &	86.3 \\
 & w/ $\mathcal{W_{CMI}}$+$\mathcal{W_{SIL}}$  &\textbf{92.6}	&\textbf{85.5}	&	\textbf{83.3}&	\textbf{85.0}&		\textbf{43.5}	&\textbf{44.7}	&\textbf{92.7} &	\textbf{86.6} \\
     
\hline
\\
\hline

    \multicolumn{2}{c}{\multirow{2}{*}{\textbf{Method}}}
& \multicolumn{2}{c}{\textbf{NUSWIDE}} & \multicolumn{2}{c}{\textbf{Fashion}}& \multicolumn{2}{c}{\textbf{CIFAR10}} 
 & \multicolumn{2}{c}{\textbf{Caltech6V20}}    \\

      &   &\textbf{ACC} &\textbf{NMI} & \textbf{ACC} & \textbf{NMI} &  \textbf{ACC} & \textbf{NMI}  &  \textbf{ACC} & \textbf{NMI} \\
     \hline
{\multirow{4}{*}{Pairwise}} & w/o $\mathcal{W}$ &  56.4 &	32.8 	&		99.4 	&98.3 		&	31.6 	&22.4 
 &			46.6 &	61.9 \\
& w/ $\mathcal{W_{CMI}}$  & 55.9 &	33.9 	&		99.4& 	98.4 	&		37.2 &	23.6 
	&		47.2 	&63.2 
\\
& w/ $\mathcal{W_{SIL}}$ & 60.9 &	32.5 	&		99.4 &	98.6 		&	35.7 	&22.7 
 	&		47.7 &	61.3 
\\
& w/ $\mathcal{W_{CMI}}$+$\mathcal{W_{SIL}}$  & 61.0 &	34.3 		&	99.5 	&98.6 	&		38.1 &	24.1 
 &			51.1 &	63.7 
\\

     \hline
{\multirow{4}{*}{\textbf{B-O (Ours)}}} & w/o $\mathcal{W}$ &  61.5 	&35.0 		&	99.3 	&98.2 	&		34.8 &	22.2 
		&	44.3 &	59.5  \\
& w/ $\mathcal{W_{CMI}}$  &62.0 &	\textbf{37.6} &			99.2 &	98.4 	&		32.8 &	23.6 	&		44.0 &	59.9 
 \\

& w/ $\mathcal{W_{SIL}}$  & 57.9 &	29.9 	&		99.4 &	98.6 		&	 37.5 &	23.0 	&		44.7 	&59.2 
 \\
 & w/ $\mathcal{W_{CMI}}$+$\mathcal{W_{SIL}}$  &	\textbf{63.1} &	37.1 	&		\textbf{99.5 }&	\textbf{98.7} 	&		\textbf{39.0} &	\textbf{24.6} &	\textbf{52.3} 	&\textbf{62.6}		
\\
     
\hline

    \end{tabular}
    \label{tab:weight_appendix}
\end{table*}

Moreover, to analyze representation degeneration in existing MVCC methods (DCP, DSMVC, DSIMVC, MFLVC, DualMVC, CVCL, SEM, ACCMVC, and DCMVC), we compare their clustering performance with that of BSV across various datasets, as shown in Fig.\ref{fig:scatter}. 
Fig.~\ref{fig:view_acc}(a) and Tab.~\ref{tab:view_improvement} reveal significant view quality discrepancies in the Caltech6V7 and Caltech5V7 datasets. 
Most MVCC methods underperform compared to BSV, exhibiting severe representation degeneration on these two datasets. Notably, while SEM generally outperforms BSV on most datasets, it still lags behind on Caltech5V7 and Caltech6V7 due to its neglect of view quality. In contrast, DWCL consistently surpasses BSV across all datasets, demonstrating its effectiveness in addressing representation degeneration. Specifically, DWCL improves ACC over BSV by 2.7\%, 3.0\%, 6.3\%, 14.9\%, 21.9\%, 29.6\%, 21.1\%, and 9.7\% across the eight datasets. This enhancement is attributed to DWCL’s consideration of both view representation and quality discrepancies, effectively tackling the issue of representation degeneration in contrastive learning.

\subsection{Ablation Studies}
\textbf{
Analysis of the B-O Contrastive Mechanism} 
To discuss the effectiveness of the B-O contrastive mechanism in detail, we use the Caltech6V7 dataset as an example. The experiments are conducted on DWCL. 
As shown in Fig.~\ref{fig:weight}(a), we select the top two best views (view 4 and view 5) and the two worst views (view 1 and view 2) to analyze the behavior of dual weights during the iterative process. 
Fig.~\ref{fig:weight}(b) illustrates that in the traditional pairwise mechanism, the dual weight $\mathcal{W}_{PW}^{4,5}$ for high-quality cross-views decreases over time, while the dual weight $\mathcal{W}_{PW}^{1,2}$ or low-quality cross-views continually increases, eventually surpassing $\mathcal{W}_{PW}^{4,5}$. This indicates that low-quality cross-views are still enhanced in the traditional pairwise contrastive mechanism. 
In contrast, our B-O contrastive mechanism sets the dual weights for low-quality cross-views, $\mathcal{W}_{B-O}^{1,2}$, to zero, preventing their enhancement during the iterative process, while the dual weight for high-quality cross-views, $\mathcal{W}_{B-O}^{4,5}$, progressively increases. More intuitively, Fig.~\ref{fig:weight}(c) shows that in the B-O contrastive mechanism, the representational capability of each view gradually improves, leading to clustering performance that exceeds that of the pairwise contrastive mechanism. These results underscore the superiority of our B-O contrastive mechanism. 
Moreover, Tab.~\ref{tab:best_view} presents the ACC and NMI results for the traditional pairwise contrastive mechanism and the B-O contrastive mechanism across eight multi-view datasets. The B-O contrastive mechanism shows significant improvements over the traditional approach in all datasets.

\textbf{
Improvements for Each View in DWCL
} 
To validate the performance improvement of our method for each view, we compare it with CVCL~\cite{chen2023deep} and SEM~\cite{xu2024self} on the Caltech5V7 and NUSWIDE datasets. 
As shown in Tab.~\ref{tab:view_improvement}, our dual-weighted strategy combined with the B-O mechanism ensures consistent and strong clustering performance across all views. 
In DWCL, the best view is determined by the highest silhouette coefficient. 
Specifically, in the Caltech5V7 dataset, View 3 is identified as the best-performing view. However, in both CVCL and SEM, the performance of View 3 deteriorates after contrastive learning, indicating representation degeneration. In contrast, our DWCL continues to enhance the performance of View 3, showing that our approach can mitigate representation degeneration.

\begin{table}
 \renewcommand\arraystretch{1.2}
 \setlength\tabcolsep{2pt}
    \centering
    \caption{
    Computation Times (in Seconds) for SEM and Our Method Across Eight Multi-View Datasets.}
    
    \begin{tabular}{l|cccccccc}
     \hline
       \textbf{Method}  & \textbf{Caltech5V7}&	\textbf{Caltech6V7}&	\textbf{Caltech6V20}& \textbf{DHA}\\
         \hline
      SEM   &  70.2&	96.1&	159.0	&10.4\\
      DCMVC & 36.4&	88.9	&143.9&	52.9\\
      \textbf{DWCL (Ours)}   &  \textbf{33.8}&	\textbf{36.2}&	\textbf{62.0}	&\textbf{8.0}\\
      \hline
      \\
      \hline
       \textbf{Method}  & \textbf{NUSWIDE}&	\textbf{CIFAR10}&	\textbf{Scene}&	\textbf{Fashion} \\
         \hline
      SEM   &237.8&	1177.8	&116.9&	262.5 \\
      DCMVC & 151.9&	- &	115.2	&207.9\\
      \textbf{DWCL (Ours)}  & \textbf{122.1}	&\textbf{340.5}&	\textbf{83.3}&	\textbf{197.8} \\
       \hline
    \end{tabular}
    \label{tab:time}
\vspace{-0.3cm}
\end{table}

\textbf{
Analysis of the Dual Weighting Strategy} 
We conduct ablation experiments under consistent conditions to assess the effectiveness of the cross-view quality weight 
$\mathcal{W}_{SI}$ and the view discrepancy weight $\mathcal{W}_{CMI}$ in contrastive loss. As shown in Tab.~\ref{tab:weight_appendix}, the best results are achieved using the dual weighting strategy, as expected. Additionally, varying levels of improvement are observed when using the view discrepancy weight and the cross-view quality weight independently. 
Specifically, compared to DWCL without any weights, the clustering accuracy of DWCL with dual weights improved significantly on the Caltech5V7, DHA, CIFAR10, and Caltech6V20 datasets by 3.3\%, 5.7\%, 4.2\%, and 8.0\%, respectively. These results demonstrate that the dual weighting strategy significantly enhances the effectiveness of feature representations across different views.

\textbf{
Impact of Different Contrastive Mechanisms and Weights} 
Tab.~\ref{tab:weight_appendix} presents the performance across eight multi-view datasets under various contrastive mechanisms and different loss weights $\mathcal{W}$. Our B-O contrastive mechanism combined with view quality weights  $\mathcal{W_{CMI}}$ + $\mathcal{W_{SIL}}$ achieves the best performance on  eight datasets. Using only our proposed weight $\mathcal{W_{SIL}}$ yields the second-best results on the Caltech5V7, Scene, Caltech6V7, and Fashion datasets when employing the B-O contrastive mechanism with the best view as the benchmark.

\begin{table} 
 \renewcommand\arraystretch{1.2}
\setlength\tabcolsep{1pt}
    \centering
        \caption{
        Clustering Performance of DWCL With and Without Reconstruction Loss Across Eight Multi-View Datasets.}         
    \begin{tabular}{l|cccccccccccc}
    \hline
    \multirow{2.5}{*}{\textbf{Method}}
& \multicolumn{2}{c}{\textbf{Caltech6V7}}& \multicolumn{2}{c}{\textbf{Caltech5V7}}& \multicolumn{2}{c}{\textbf{Fashion}}  &\multicolumn{2}{c}{\textbf{Scene}}
    \\
\cline{2-9}

          &\textbf{ACC}&\textbf{NMI}&  \textbf{ACC} & \textbf{NMI} &  \textbf{ACC} & \textbf{NMI} & \textbf{ACC} & \textbf{NMI}   \\
     \hline
     DWCL w/ $\mathcal{R}$ & 92.3	&86.2&	92.6&	85.5	&\textbf{99.5} &	\textbf{98.6} &	43.5	&44.7	
  \\
     DWCL w/o $\mathcal{R}$ & \textbf{92.5}	&\textbf{86.6}	&	\textbf{93.8}	&\textbf{88.6}	&	99.4	&98.4&	\textbf{44.9}&	\textbf{45.6}
   \\
\hline
\\
\hline
    \multirow{2.5}{*}{\textbf{Method}}
& \multicolumn{2}{c}{\textbf{NUSWIDE}}& \multicolumn{2}{c}{\textbf{DHA}}& \multicolumn{2}{c}{\textbf{Caltech6V20}}  &\multicolumn{2}{c}{\textbf{CIFAR10}}
    \\
\cline{2-9}

          &\textbf{ACC}&\textbf{NMI}&  \textbf{ACC} & \textbf{NMI} &  \textbf{ACC} & \textbf{NMI} & \textbf{ACC} & \textbf{NMI}   \\
     \hline
     DWCL w/ $\mathcal{R}$ &  \textbf{62.2}	&\textbf{37.2} & \textbf{83.3}&	\textbf{85.0} &\textbf{52.4}&	\textbf{63.7}
&\textbf{39.0}	&\textbf{24.6}
  \\
     DWCL w/o $\mathcal{R}$ & 56.7&	24.9&		 82.0	&82.8&	49.1	&58.8	& 28.2&	15.4
        \\
\hline
    \end{tabular}
    \label{tab:reconstruction}
    \vspace{-0.3cm}
\end{table}

\begin{table}
\setlength\tabcolsep{2pt}
 \renewcommand\arraystretch{1}
    \centering
        \caption{
        Classification Performance of DWCL.}      
    \begin{tabular}{l|cccccc}
    \hline
    \multirow{2.5}{*}{\textbf{Method}}
& \multicolumn{3}{c}{\textbf{DHA}}& \multicolumn{3}{c}{\textbf{NUSWIDE}}\\
\cline{2-7}
  
          &\textbf{ACC}&\textbf{Recall}& \textbf{F1-Score} & \textbf{ACC} & \textbf{Recall}& \textbf{F1-Score}\\
     \hline
    DCP& 64.4& 66.2&61.9& 66.4&66.9&62.4\\
    DSMVC&66.1&67.1&62.9&65.6&65.7&61.8\\
    DSIMVC&73.6&66.1&71.0& 66.1 &62.4&	64.7\\
    MFLVC&	71.3	&72.8&68.5	&70.2  &70.3	&{70.4}	\\
    DualMVC&73.6	& 75.6  &	69.7&69.9  &70.0	&70.0	\\
    CVCL&76.9	& {}{78.9}  &71.9	& 62.5 &62.6	&62.2	\\
    SEM&{}{77.8}	& 78.3  &{}{77.5}	& {}{70.3} &	{}{70.3}&70.4	\\
    ACCMVC &   76.1	&77.2&	75.0	&63.1&	63.0	&63.8\\
    DCMVC &   73.5	&75.4&	71.6&	71.2&	71.3	&71.3\\
    \textbf{DWCL (Ours)}&\textbf{81.2}	&\textbf{82.5}   &\textbf{80.2}	& \textbf{73.2} &\textbf{73.1}	&\textbf{73.3}	\\
     
\hline
    \end{tabular}
    \label{tab:classification}
    \vspace{-0.3cm}
    
\end{table}

\textbf{Efficiency of Our DWCL} 
As shown in Tab.~\ref{tab:time}, we compare the computational time of our DWCL method with DCMVC~\cite{cui2024dual} and SEM~\cite{xu2024self}, a state-of-the-art approach in self-weighted pairwise contrastive learning. 
The time for a single iteration of contrastive learning is evaluated across multiple datasets, ensuring each iteration uses the same number of training epochs for consistency. 
Our DWCL method consistently outperforms SEM in terms of speed in seconds, being 2 to 3 times faster while maintaining superior performance. 
This substantial reduction in computational time demonstrates that our DWCL approach is not only more efficient but also significantly reduces computational costs compared to the self-weighted pairwise contrastive mechanism.

\begin{figure}
    \centering
    \includegraphics[width=0.9\linewidth]{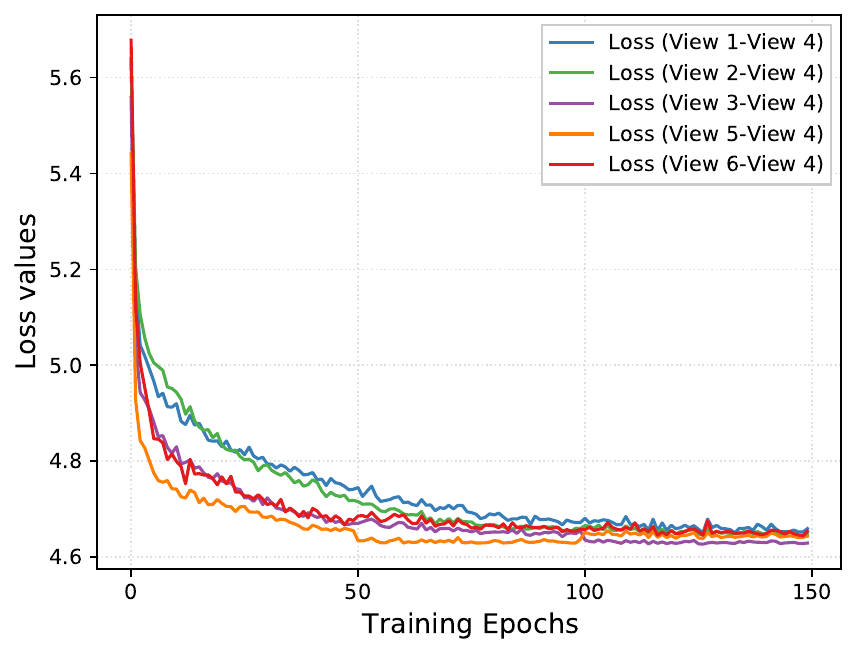}
    \Description{}
    \caption{
    Convergence Results of the Best View Compared to Each View Obtained by DWCL on Caltech6V7.}
    \label{convergence}
    \vspace{-0.3cm}
\end{figure}

\begin{figure}
\centering 
\subfloat[ACC]{
\label{fig:5v_acc_appendix}
\includegraphics[width=0.24\textwidth]{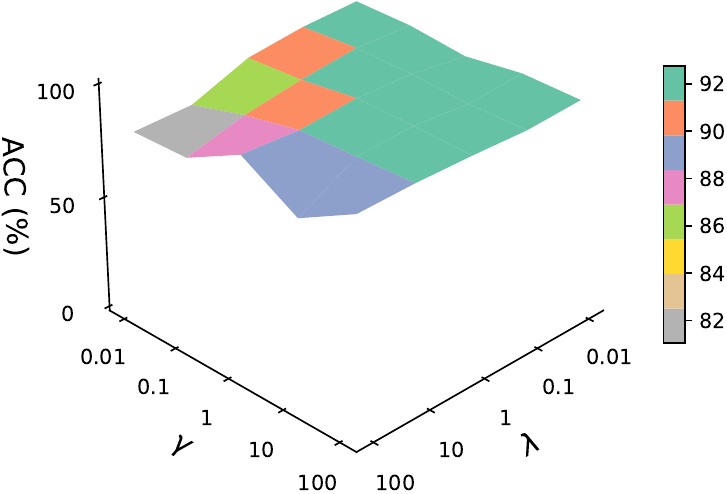}}
\subfloat[NMI]{
\label{fig:5v_nmi_appendix}
\includegraphics[width=0.24\textwidth]{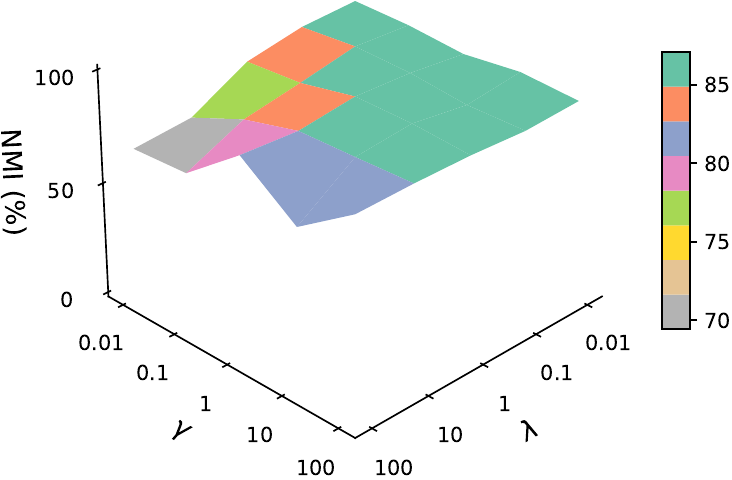}}
\caption{
Results of ACC and NMI for Different Combinations of $\lambda$ and $\gamma$ on Caltech6V7.
}
    \Description{}
    \vspace{-0.45cm}
\label{fig:parameter_appendix1}
\end{figure}

\begin{figure}[!h]
\centering 
\subfloat[ACC]{
\label{fig:5v_acc_nus}
\includegraphics[width=0.24\textwidth]{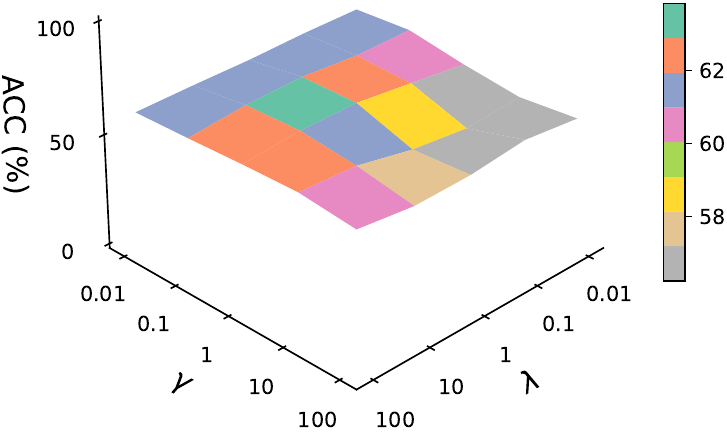}}
\subfloat[NMI]{
\label{fig:5v_nmi_nus}
\includegraphics[width=0.24\textwidth]{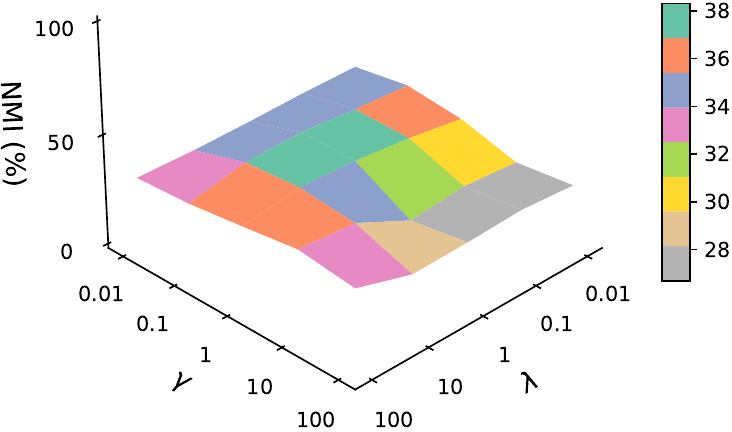}}
\caption{
Results of ACC and NMI for Different Combinations of $\lambda$ and $\gamma$ on NUSWIDE.}
    \Description{}

\label{fig:parameter_appendix2}

\end{figure}

\begin{figure*}[t]
\centering 
\subfloat[MFLVC]{
\label{fig:MFLVC}
\includegraphics[width=0.33\textwidth]{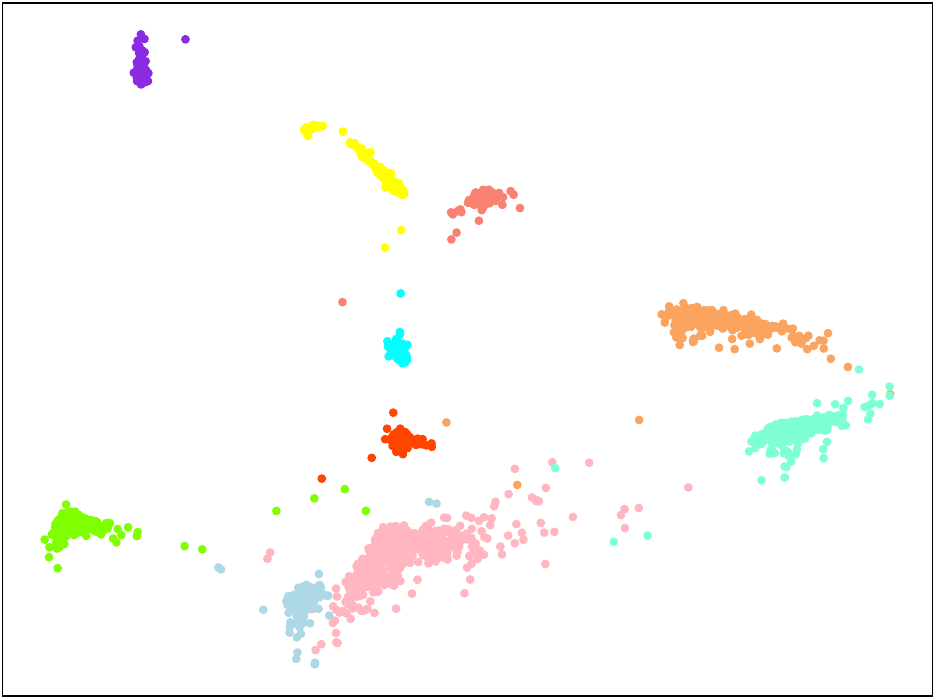}}
\subfloat[SEM]{
\label{fig:SEM}
\includegraphics[width=0.33\textwidth]{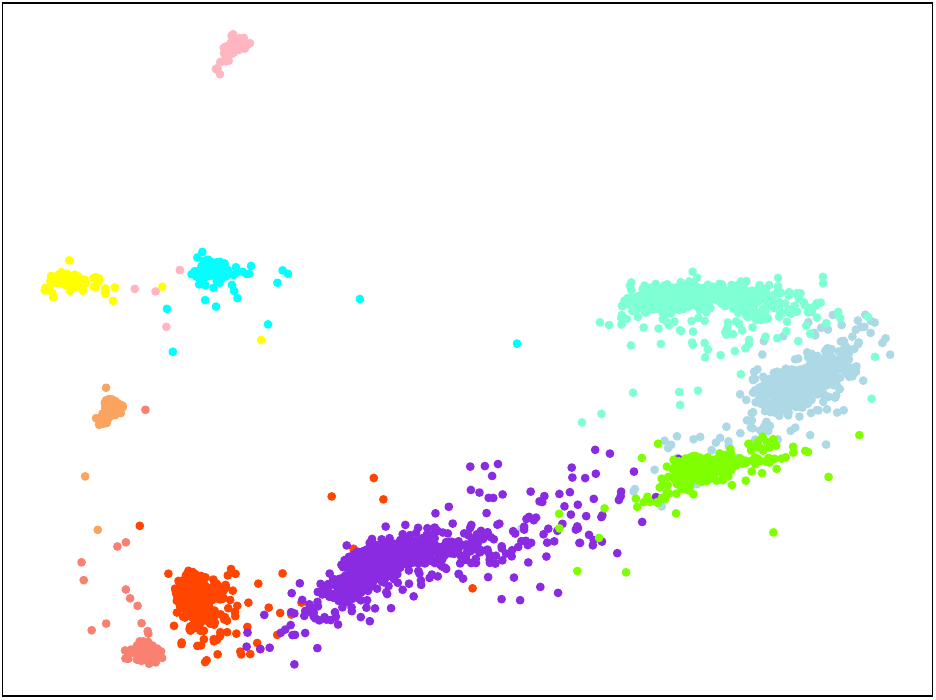}}
\subfloat[DWCL]{
\label{fig:xxxx}
\includegraphics[width=0.33\textwidth]{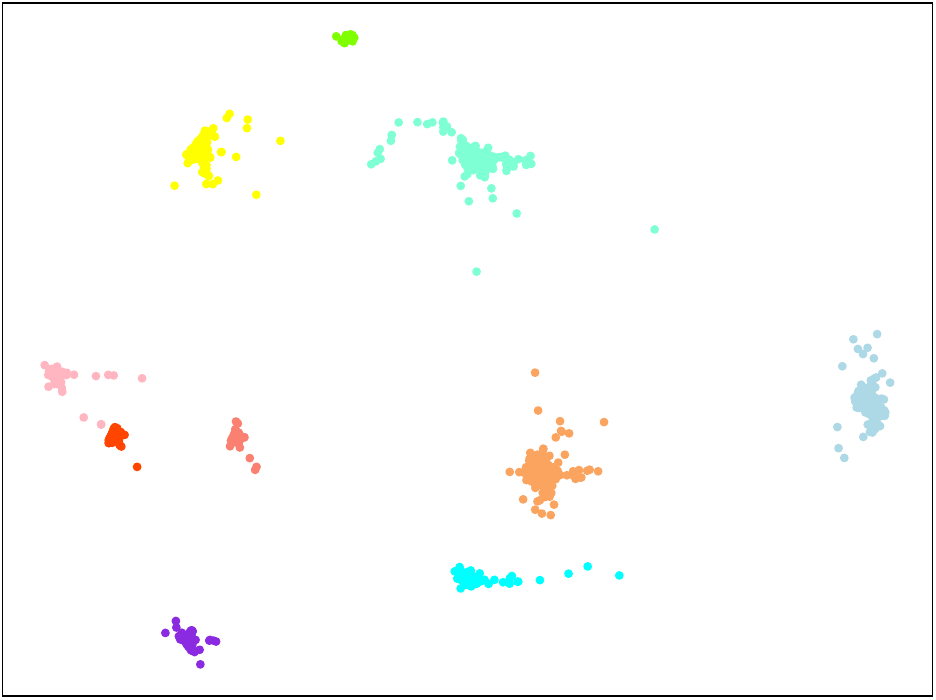}}
\caption{
Visual Comparison of Two Representative Multi-View Contrastive Learning Methods on the Fashion Dataset.}
    \Description{}

\label{fig:visual_appendix}
\end{figure*}

\begin{figure}

	\centering  
	\subfloat[Epoch 0]{
		\label{fashion_Epoch0}		\includegraphics[width=0.45\linewidth]{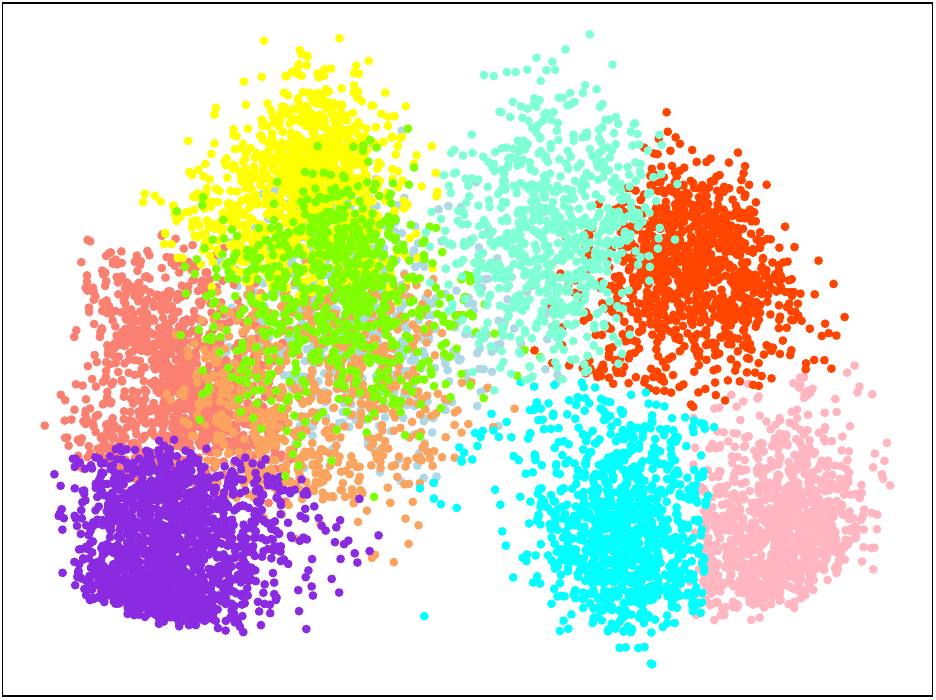}}
	\quad 
	\subfloat[Epoch 100]{
		\label{fashion_Epoch100}
\includegraphics[width=0.45\linewidth]{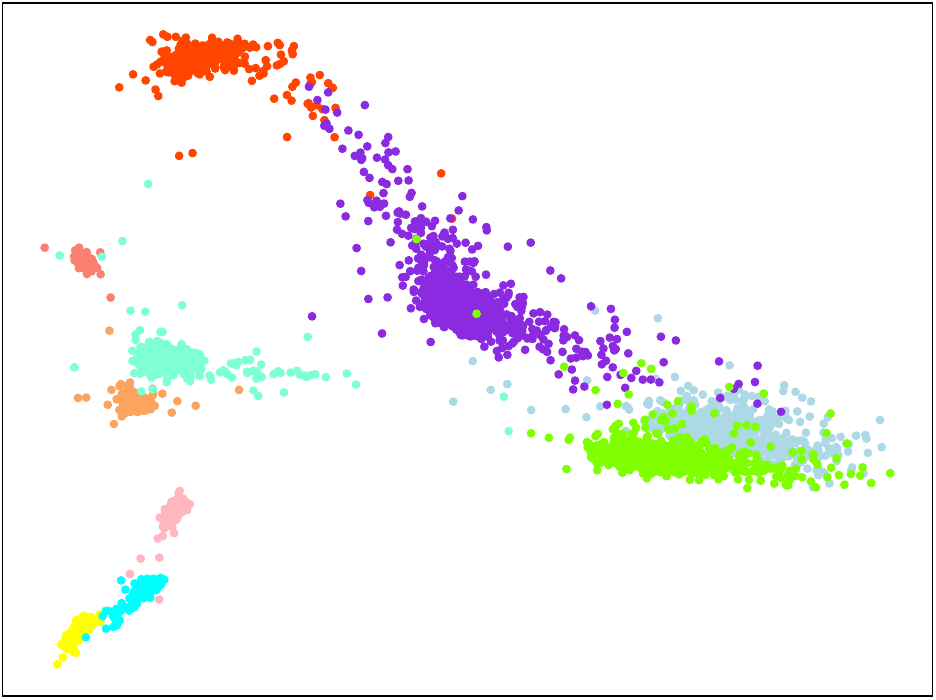}}

	\subfloat[Epoch 200]{
		\label{fashion_Epoch200}
\includegraphics[width=0.45\linewidth]{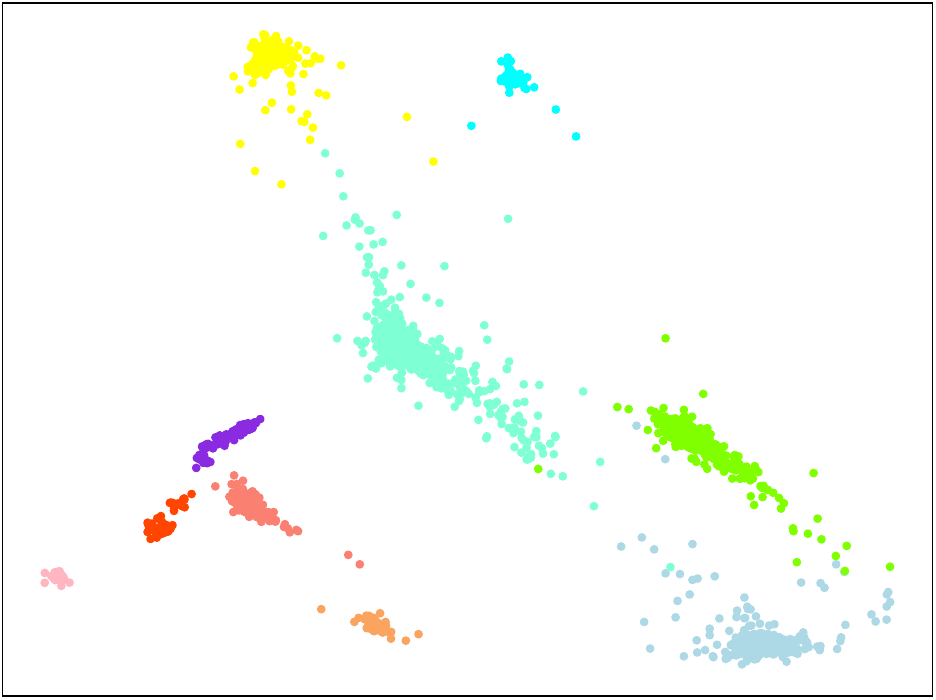}}
	\quad
	\subfloat[Epoch 300]{
		\label{fashion_Epoch300}
\includegraphics[width=0.45\linewidth]{visual_fashion_300.pdf}}
	\caption{
    Visualization of DWCL in the Contrastive Learning Process on the Fashion Dataset.}
	\label{visual_train}
\end{figure}

\textbf{Analysis of Reconstruction Regularization in DWCL} 
Tab.~\ref{tab:reconstruction} analyzes the impact of reconstruction loss $\mathcal{L}_R$ on the performance of the DWCL method across eight multi-view datasets. Notably, our findings show that DWCL without reconstruction loss outperforms the version with it on the Caltech5V7 dataset. Additionally, similar competitive results are observed on other datasets, including Caltech6V7, Fashion, and Scene. These results suggest that DWCL can achieve robust and effective clustering performance even without reconstruction loss, highlighting its strength and ability to deliver strong results independently of this component.

\textbf{Performance of DWCL in Downstream Tasks} 
To further validate the quality of the representations learned by DWCL, we evaluate its effectiveness by applying the features extracted through DWCL to a downstream classification task, utilizing the SVM~\cite{cortes1995support} linear classification algorithm. 
For this experiment, we selected two datasets: DHA and NUSWIDE, which were divided into training and testing subsets with a 70\% to 30\% split, respectively. The results, presented in Tab.~\ref{tab:classification}, clearly demonstrate that DWCL outperforms competing methods across three critical evaluation metrics: accuracy (ACC), recall, and F1-score. 
These results strongly suggest that DWCL is not only capable of learning high-quality, discriminative feature representations but also excels in supervised learning tasks, making it a highly effective approach for downstream classification applications.

\textbf{Training Analysis} 
We investigate the sensitivity of DWCL to the weights of the reconstruction loss ($\lambda$) and contrastive loss ($\gamma$) on the Caltech6V7 and NUSWIDE datasets. The values for $\lambda$ and $\gamma$ are chosen from the set [0.01, 0.1, 1, 10, 100], covering a wide range of possible configurations. Fig.~\ref{fig:parameter_appendix1} and \ref{fig:parameter_appendix2} present the clustering performance of DWCL, evaluated in terms of accuracy (ACC) and normalized mutual information (NMI), for various combinations of $\lambda$ and $\gamma$. The experimental results show that DWCL exhibits strong robustness across the majority of parameter settings, maintaining consistent and reliable performance regardless of the specific values chosen for $\lambda$ and $\gamma$.
This behavior is observed in both small and large datasets, suggesting that DWCL is not sensitive to the choice of these hyper-parameters and can deliver effective clustering results across a wide range of conditions.

\textbf{Parameter Sensitivity} 
We investigate the sensitivity of the DWCL method to the reconstruction loss weight $\lambda$ and the contrastive loss weight $\gamma$. We assess how variations in these parameters impact clustering performance on two benchmark datasets: Caltech6V7 and NUSWIDE. The weights for $\lambda$ and $\gamma$ are selected from the set [0.01, 0.1, 1, 10, 100].
Clustering results, measured by Accuracy (ACC) and Normalized Mutual Information (NMI), are shown in Fig.~\ref{fig:parameter_appendix1} and Fig.~\ref{fig:parameter_appendix2}. Our analysis indicates that DWCL maintains robust clustering performance with minimal sensitivity to changes in $\lambda$ and $\gamma$. It consistently achieves strong results across both smaller datasets like Caltech6V7 and larger ones such as NUSWIDE, highlighting its effectiveness and generalizability.

\textbf{Visualization} 
To illustrate the effectiveness of DWCL, we use the Fashion dataset as a case study. Fig.~\ref{visual_train} shows the evolution of clustering results achieved by DWCL during the contrastive learning process. As training progresses, the clustering structure becomes increasingly distinct, demonstrating the method's ability to enhance class separation over time.
Additionally, Fig.~\ref{fig:visual_appendix} visually compares DWCL's performance with two multi-view clustering methods, MFLVC~\cite{xu2022multi} and SEM~\cite{xu2024self}. The comparison clearly indicates that DWCL outperforms both methods, producing the most well-defined and separated clusters. This visual evidence underscores DWCL’s superior capability in learning discriminative features and improving clustering quality relative to existing approaches.

\section{Conclusion}
In this paper, we propose Dual-Weighted Contrastive Learning (DWCL) to effectively and efficiently address representation degeneration in multi-view contrastive clustering. Specifically, DWCL introduces the innovative Best-Other multi-view contrastive mechanism and a view quality weight to optimize contrastive learning. By eliminating unreliable cross-views and fully leveraging the representation capabilities of high-quality views, DWCL achieves robust and efficient multi-view clustering. Theoretical analysis and extensive experiments confirm the efficiency and the effectiveness of our approach. 
While our B-O contrastive mechanism shows promise, the best view may not always be the ideal choice in every scenario, which could impact clustering performance in certain cases. Future work will aim to develop more precise metrics for evaluating view quality, allowing DWCL to adapt to a wider range of scenarios and enhancing its robustness in diverse multi-view contexts.



\end{document}